\documentclass{article}
\usepackage[left=2.5cm,right=2.5cm,top=2cm,bottom=2cm]{geometry}
\usepackage{float}
\usepackage{amsmath}
\usepackage[english]{babel}
\usepackage[autostyle, english = american]{csquotes}
\MakeOuterQuote{"}

\usepackage[utf8]{inputenc}
\usepackage{amsfonts,amssymb}
\usepackage{graphicx} 
\usepackage[square,numbers]{natbib}
\usepackage{subcaption}
\usepackage{hyperref}
\usepackage{booktabs}
\usepackage{authblk}
\usepackage{multirow}
\usepackage{xcolor}
\usepackage{listings}
\usepackage[toc,page]{appendix}

\newcommand{\Rb}{\mathbb{R}}

\NewDocumentCommand{\codeword}{v}{%
\texttt{\textcolor{blue}{#1}}%
}

\begin{document}

\title{LEMs: A Primer On Large Execution Models}

\author{Rémi Genet\textsuperscript{\textsection}\footnote{DRM, Université Paris Dauphine - PSL, Pl. du Maréchal de Lattre de Tassigny, 75016 Paris, France. E-mail address: remi.genet@dauphine.psl.eu} , Hugo Inzirillo\textsuperscript{\textsection}\footnote{ESG-UQAM, Montreal, QC, Canada. E-mail address: inzirillo.hugo@euqam.ca}}
\begingroup\renewcommand\thefootnote{\textsection}
\footnotetext{These authors contributed equally.}
\endgroup

\maketitle

\begin{abstract}
This paper introduces Large Execution Models (LEMs), a novel deep learning framework that extends transformer-based architectures to address complex execution problems with flexible time boundaries and multiple execution constraints. Building upon recent advances in neural VWAP execution strategies, LEMs generalize the approach from fixed-duration orders to scenarios where execution duration is bounded between minimum and maximum time horizons, similar to share buyback contract structures. The proposed architecture decouples market information processing from execution allocation decisions: a common feature extraction pipeline using Temporal Kolmogorov-Arnold Networks (TKANs), Variable Selection Networks (VSNs), and multi-head attention mechanisms processes market data to create informational context, while independent allocation networks handle the specific execution logic for different scenarios (fixed quantity vs. fixed notional, buy vs. sell orders). This architectural separation enables a unified model to handle diverse execution objectives while leveraging shared market understanding across scenarios. Through comprehensive empirical evaluation on intraday cryptocurrency markets and multi-day equity trading using DOW Jones constituents, we demonstrate that LEMs achieve superior execution performance compared to traditional benchmarks by dynamically optimizing execution paths within flexible time constraints. The unified model architecture enables deployment across different execution scenarios (buy/sell orders, varying duration boundaries, volume/notional targets) through a single framework, providing significant operational advantages over asset-specific approaches. Our results suggest that the deep learning methodologies developed for standard VWAP execution can be successfully extended to more complex execution problems, offering a scalable solution for institutional trading desks facing diverse execution requirements.

\begin{figure}[H]
    \centering
    \includegraphics[width=0.4\linewidth]{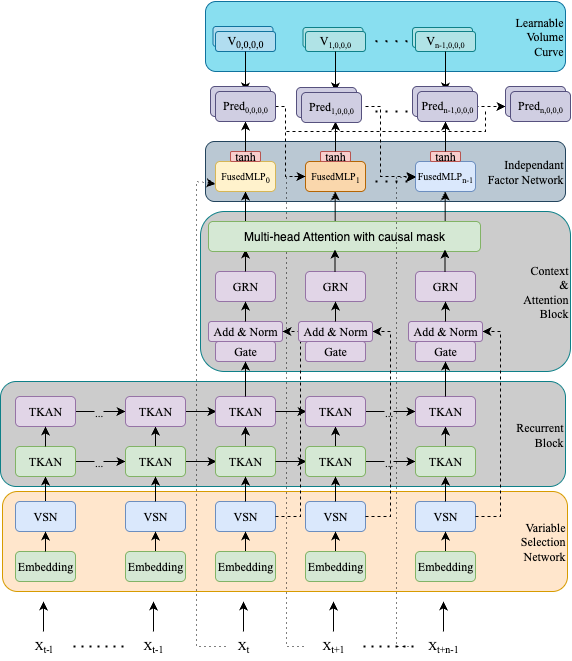}
    \caption{Large Execution Models (LEMs)} 
    \label{fig:tkan}
\end{figure}
\end{abstract}

\section{Introduction}
The optimization of transaction execution is a key challenge in quantitative finance, where determining "when" and "how" to trade in the markets is essential to minimize costs and maximize performance. A fundamental aspect of this problem concerns the market impact generated by order execution, particularly for large orders. Market impact refers to the phenomenon whereby executing an order influences the asset’s price, generally creating an adverse movement for the initiator of the transaction and thus an additional implicit cost. Early studies have shown that this impact is a concave function of order size, and this impact has since frequently been decomposed into temporary and permanent components, or represented as impact that dissipates over time.

\medskip

Researchers proposed numerous models to quantify these impacts \cite{Loeb1983,Hasbrouck1991,Alfonsi2010,Bouchaud2004,Hey2025Concave}, but two determining factors consistently emerge: the proportion that the order represents relative to the volumes usually traded in the market, and the asset’s intrinsic volatility. In the context of cryptocurrency markets, characterized by generally lower liquidity and higher volatility than traditional financial markets, market impact becomes a particularly critical issue. Rather than seeking instant execution, it therefore appears pertinent to extend the execution duration of the order so that its participation in the market volume becomes negligible, thus reducing its impact. In practice, the trader, or the broker in charge of execution, receives the initial order, which he or she divides into several sub-orders; the aggregate order is then referred to as a metaorder. However, this approach immediately raises an evaluation question: how can one objectively measure the execution quality of an order over a period of time?

\medskip

It is in this context that the VWAP has established itself as an essential reference. The VWAP represents the average price at which the asset actually traded during the period considered, calculated by weighting each transaction price by its corresponding volume. This measure has a unique and particularly relevant mathematical property: over any horizon, the sum of the deviations between individual transaction prices and the VWAP, weighted by the corresponding volumes, necessarily cancels out. This property makes it a natural benchmark for evaluating execution performance. This use of the VWAP as an evaluation metric gave rise to a specific order type: the VWAP order, which today ranks among the most commonly issued execution instructions by financial institutions. Unlike a limit order or a standard market order, the VWAP order executes over a defined time period and is distinguished by its performance objective: to minimize the gap between the average execution price and the market VWAP over that period. This formulation, although seemingly simple, conceals considerable complexity, as it implies the simultaneous consideration of two interdependent variables: price and volume.

\medskip

The evolution of academic literature on VWAP strategies reflects different approaches to this complexity. Early works, such as those by \citet{konishi2002optimal} and \citet{Culoch2007}, explicitly integrated the relationship between volatility and volume into their allocation models. This approach recognized the well-documented empirical finance stylized fact that periods of high volatility generally coincide with high trading volumes. However, a subsequent trend developed in the literature, favoring an exclusive focus on modeling volumes at the expense of explicitly accounting for price–volume interaction. This approach, notably illustrated by the works of \citet{LeFol2006,LeFol2012}, rests on a valid theoretical reasoning: if one could perfectly predict the future volume curve, it would be possible to execute a perfect VWAP by allocating orders proportionally to this curve. This simplification allowed the integration of more sophisticated models such as Self-Exciting Threshold AutoRegressive (SETAR) proposed by \cite{tong1990non}, which apply different autoregressive models according to thresholds determined by the data themselves. The use of this type of modeling is particularly relevant because the observation of volume dynamics quickly reveals that a purely linear relationship would be insufficient to capture the observable regime changes. Another major advantage of this simplification is that it greatly facilitates the development of dynamic allocation models, enabling real-time adjustment of order allocation, a particularly desirable property for execution, which by its nature is a dynamic problem. This formalization becomes significantly more complex when one attempts to model simultaneously the interdependent dynamics of prices and volumes.

\medskip

Nevertheless, this simplification rests on two particularly problematic implicit assumptions in the context of cryptocurrency markets: the ability to predict future volumes accurately, and the statistical independence between prediction errors and price volatility. The first assumption runs up against the intrinsic unpredictability of financial markets, exacerbated in the crypto ecosystem by its relative youth and sensitivity to exogenous events. The second overlooks a robust empirical fact: volumes and volatility exhibit a significant positive correlation, particularly during major market moves, precisely the moments when execution quality becomes most critical. The execution of large orders in financial markets has long been recognized as a fundamental challenge in quantitative finance, with implications extending far beyond simple transaction cost minimization. Traditional approaches to optimal execution, rooted in the seminal work of Almgren and Chriss and subsequent optimal control methodologies, have provided valuable theoretical frameworks but often struggle with the practical complexities of modern market microstructure and the diverse execution requirements encountered in institutional trading environments. The landscape of execution strategies has undergone significant transformation with the emergence of deep learning techniques in financial applications. Recent work has demonstrated that neural networks can effectively capture complex temporal dependencies and market patterns in Volume Weighted Average Price (VWAP) execution scenarios \cite{genet2025sigtransvwap}, achieving substantial improvements over traditional volume-based approaches. These advances suggest that the sophisticated pattern recognition capabilities of modern machine learning architectures can be successfully applied to execution problems, opening new avenues for addressing previously intractable optimization challenges.

\medskip

However, the ideas developed for standard VWAP execution can be transposed to more sophisticated requirements encountered in institutional markets. Many real-world execution scenarios involve flexible time boundaries, mixed objectives combining volume and notional constraints, and dynamic stopping criteria that differ fundamentally from fixed-schedule execution frameworks. Consider, for instance, share buyback contracts where banks face optimal stopping problems with floating maturities and notional amounts \cite{baldacci2024dispensing, gueant2017optimal}. These contracts typically involve execution windows bounded between minimum and maximum durations, coupled with flexible notional targets that can be adjusted based on market conditions and execution performance.

\medskip

The mathematical complexity of such problems has traditionally led practitioners to rely on optimal control methods. Despite their well-documented limitations in high-dimensional settings \cite{jaimungal2017optimal}. As noted by Baldacci et al. \cite{baldacci2024dispensing}, traditional stochastic optimal control approaches encounter three significant challenges: the curse of dimensionality when dealing with complex contracts, the difficulty of selecting appropriate risk parameters, and incompatibility with standard pricing and hedging frameworks used in practice. These limitations have motivated the exploration of alternative approaches, including the heuristic optimization strategies that have shown promise in share buyback applications. During the recent year, the evolution of transformer architectures \cite{vaswani2017attention} in financial time series analysis \cite{lim2021temporal} has demonstrated remarkable success in capturing both local and global temporal dependencies using various extensions \cite{genet2024tkat}. The "self-attention" mechanism's ability to model complex interactions across different time horizons, combined with the theoretical foundations provided by Kolmogorov-Arnold Networks \cite{liu2024kan}, offers a powerful framework for addressing execution problems that require sophisticated temporal modeling. Recent work has shown that such architectures can achieve superior performance in VWAP execution tasks by effectively balancing short-term market dynamics with longer-term strategic considerations.

\bigskip

Codes are available at \href{https://github.com/remigenet/LargeExecutionModels}{LEMs repository} and can be installed using the following command: \codeword{pip install lems}. Data are accessible if the reader wishes to reproduce our experiments using the GitHub link provided above.

\subsection{Problem Formulation}

To formalize the execution problem addressed in this paper, consider a trader who must execute an order within a flexible time window $[T_{\min}, T_{\max}]$ where $0 < T_{\min} \leq T_{\max}$. Unlike traditional VWAP execution with fixed horizons, the trader has the flexibility to choose the optimal stopping time $\tau \in [T_{\min}, T_{\max}]$ to maximize execution performance relative to a benchmark. Let $S_t$ denote the asset price at time $t$, and $V_t$ the market volume. The trader's objective may involve either:
\begin{itemize}
    \item \textbf{Fixed Quantity Execution}: Execute a predetermined number of shares $Q$ over the chosen horizon $\tau$
    \item \textbf{Fixed Notional Execution}: Spend a predetermined amount $F$ over the chosen horizon $\tau$
\end{itemize}

For the fixed quantity case, let $q_t$ denote the number of shares executed at time $t$, subject to the constraint $\sum_{t=1}^{\tau} q_t = Q$. The execution price achieved is:
\begin{equation}
P_{\text{exec}} = \frac{\sum_{t=1}^{\tau} S_t q_t}{Q}
\end{equation}

The market VWAP over the same period is:
\begin{equation}
\text{VWAP}_{\tau} = \frac{\sum_{t=1}^{\tau} S_t V_t}{\sum_{t=1}^{\tau} V_t}
\end{equation}
For the fixed notional case, given a predetermined expenditure $F$, the trader executes shares $q_t$ at each time step subject to $\sum_{t=1}^{\tau} S_t q_t = F$, aiming to optimize execution performance relative to the benchmark VWAP over the chosen horizon.

The optimization problem becomes:
\begin{equation}
\max_{\{q_t\}_{t=1}^{T_{\max}}, \tau \in [T_{\min}, T_{\max}]} \mathbb{E}\left[ U\left(\frac{P_{\text{exec}}}{\text{VWAP}_{\tau}}, \tau\right) \right]
\end{equation}
where $U(\cdot, \cdot)$ represents a utility function that captures execution quality relative to the benchmark and timing preferences.

\medskip  

The challenge lies in the fact that both future prices $\{S_t\}_{t>\text{current}}$ and volumes $\{V_t\}_{t>\text{current}}$ are unknown, requiring predictive models that can capture the complex dependencies between price dynamics, volume patterns, and optimal execution decisions. Moreover, the discrete nature of trading decisions and the need to respect various market constraints (minimum trade sizes, market impact considerations, etc.) add further complexity to the optimization problem.

\subsection{Proposed Approach}

This paper introduces Large Execution Models (LEMs), a unified deep learning framework designed to address the complex optimization problem outlined above. The architecture draws inspiration from recent advances in transformer-based VWAP execution \cite{genet2025sigtransvwap} while extending the methodology to handle flexible time boundaries and mixed execution objectives. The key architectural innovation lies in the decoupling of market information processing from allocation decision logic. The LEMs framework employs a common feature extraction pipeline that combines Temporal Kolmogorov-Arnold Networks (TKANs) for capturing complex temporal dependencies \cite{genet2024tkat}, Variable Selection Networks for adaptive feature importance weighting, and multi-head attention mechanisms for modeling long-range market interactions. This shared processing creates a rich informational context from market data that serves as input to independent allocation networks, each specialized for different execution scenarios (fixed quantity, fixed notional, buy orders, sell orders).This decoupling enables the model to leverage common market understanding across different execution modes while maintaining specialized logic for each scenario's specific requirements. The allocation networks remain relatively simple, focusing on the execution speed decisions, while the complex market feature processing is handled by the shared components.

\medskip

This paper demonstrates that deep learning frameworks naturally accommodate the exact problem formulations encountered in institutional execution, making it straightforward to represent complex optimization objectives within neural architectures. While this work focuses on near-vanilla contracts with variations around standard VWAP execution, incorporating more specific contract specifications encountered in practice represents simply additional operational steps over the proposed framework, rather than fundamental architectural changes.The key insight underlying our approach is that sophisticated temporal modeling capabilities can be shared across execution scenarios, while scenario-specific allocation logic can be handled by independent networks that operate on the common informational context. This allows a single model deployment to handle diverse execution requirements while leveraging cross-scenario learning to improve overall performance.

\section{Architecture}
We proposed an architecture based on several sub layers and simultaneously leverages the Kolmogorov Arnold networks (KANs) proposed by Liu et al. \cite{liu2024kan}. \emph{LEMs} is tailored for multi-step execution strategies across multiple market scenarios. The global objective of this model is to transforms inputs into structured allocation decisions. \emph{LEMs} have been designed to perform such a task using different modules which will be detailed in further section.

\medskip

\subsection{Model Architecture Overview}

The LEMs architecture is composed of two primary components that work in tandem to generate optimal execution strategies. The first component, the Decision Context Generation Block, processes current and historical market data to create rich contextual representations, drawing architectural inspiration from Temporal Kolmogorov-Arnold Networks (TKAN) and Temporal Fusion Transformers (TFT). This component provides the informational foundation for execution decisions. The second component, the Execution Decision Block, utilizes the generated context alongside its own previous decisions to make step-wise allocation choices throughout the execution timeline. Each execution step employs a dedicated decision module that adapts to the temporal progression of the execution strategy.

\medskip

The model processes input tensor $X \in \mathbb{R}^{B \times T \times D}$, where $B$ is the batch size, $T$ is the sequence length covering both lookback and forward execution periods, and $D$ is the number of input features including market data, trading constraints, and execution parameters.

\subsection{Decision Context Generation Block}

\subsubsection{Embedding Layer}
The model begins with an Embedding layer that transforms each feature dimension of the input independently into a higher-dimensional embedding space. Each slice $X^{(i)} \in \mathbb{R}^{B \times T \times 1}$ along the feature axis is passed through a corresponding dense layer $f_i: \mathbb{R} \rightarrow \mathbb{R}^H$, where $H$ is the specified number of hidden units, i.e., embedding dimension. This transforms each feature independently into an embedding $E^{(i)} \in \mathbb{R}^{B \times T \times H}$. All these embeddings are then stacked along a new axis, producing the final output tensor $E \in \mathbb{R}^{B \times T \times H \times D}$. Thus, the layer acts as a learnable embedding mechanism that projects each scalar feature channel into its own embedding space of dimension $H$, while maintaining the temporal structure of the data.This vector of embeddings serves as the starting point, passed through Variable Selection Networks (VSN) enhanced with the use of Gated Residual Networks (GRNs).

\subsubsection{Variable Selection Networks}

The relationships between dependent random vectors is a key issue. Gated residual networks, the key component in the Variable Selection Networks (VSN) offer an efficient and flexible way of modelling complex relationships in time series. They allow to control the flow of information and facilitate the learning tasks. They are particularly useful in areas where non-linear interactions and long-term dependencies are crucial. In our model we use the Gated Residual Networks proposed by \cite{lim2021temporal}, and we kept the same architecture. However, we remove the need for the context.

\begin{figure}[H]
    \centering
    \includegraphics[width=0.4\linewidth]{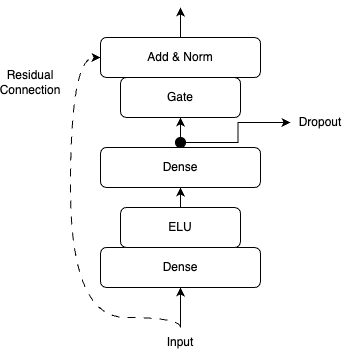}
    \caption{Gated Residual Networks (GRN) \cite{lim2021temporal}} 
    \label{fig:grn}
\end{figure}

\begin{align}
\text{GRN}_\omega\left(x \right) &=\text{LayerNorm}\left(x + \text{GLU}_\omega(\eta_1) \right), \\
    \eta_1 &= W_{1, \omega}~\eta_2 + b_{1, _\omega},  \label{eqn:grn_step}\\
    \eta_2 &= \text{ELU}\left( W_{2, \omega}~x + b_{2, _\omega} \right).  \label{eqn:grn_base}
\end{align}

In this context, $\text{ELU}$ designates the Exponential Linear Unit activation function \cite{clevert2015fast}, while $\eta_1 \in \mathbb{R}^{d_{model}}$ and $\eta_2 \in \mathbb{R}^{d_{model}}$ represent intermediate layers. The standard layer normalization $\text{LayerNorm}$ is that described in \cite{ba2016layer}, and $\omega$ is an index used to indicate weight sharing. When the expression $W_{2, \omega} x + b_{2, \omega}$ is largely positive, ELU activation works as an identity function. On the other hand, when this expression is largely negative, ELU activation produces a constant output, thus behaving like a linear layer. We use Gated Linear Units (GLUs) \cite{dauphin2017language}. They provide the flexibility to suppress any parts of the architecture that are not required for a given dataset. Letting $\gamma \in \mathbb{R}^{d_{model}}$ be the input, the GLU then takes the form
\begin{align}
 \text{GLU}_\omega(\gamma) & =  \sigma(W_{4, \omega}~\gamma + b_{4, \omega}) \odot (W_{5, \omega}~\gamma + b_{5, \omega} ),
\label{eqn:component_gate}
\end{align}
where $\sigma(.)$ is the sigmoid activation function; 
$W(.) \in \mathbb{R}^{d_{model}\times d_{model}}$, $b(.) \in \mathbb{R}^{d_{model}}$ are weights and intercepts, respectively; $\odot$ denotes the element-wise Hadamard product, and $d_{model}$ is the hidden state size (common across TFT).
GLU allows TFT to control the extent to which the GRN contributes to the original input $x$ -- potentially skipping over the layer entirely if necessary, as the GLU outputs could be all close to 0 in order to suppress the nonlinear contribution of the input variable to forecasting.

\begin{figure}[H]
    \centering
    \includegraphics[width=0.5\linewidth]{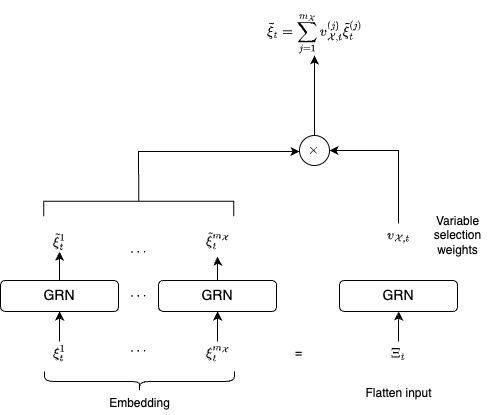}
    \caption{Variable Selection Networks (VSN)} 
    \label{fig:vsn}
\end{figure}

VSN will help model performance via the use of learning capacity only for the most salient covariates. Figure \ref{fig:vsn} described the different steps in the VSN. We define ${\xi}_{t}^{(j)}  \in \mathbb{R}^{d_{model}}$  the transformed input of the $j$-th variable at time $t$, with ${\Xi}_t = \left[ {\xi}_{t}^{(1)^T}, \dots, {\xi}_{t}^{(m_\chi)^T}  \right]^T$, the flattened vector of all past inputs at time $t$. Variable selection weights, denoted $v_{\mathcal{X},t}$, are generated by feeding ${\Xi}_t$ through a GRN and followed by a Softmax layer: 
\begin{align}
{v}_{\chi_t} = \text{Softmax}\left(\text{GRN}_{v_{\chi}}({\Xi}_t)\right),& 
\label{eq:vsn}
\end{align}
where ${v}_{\chi_t} \in \mathbb{R}^{m_\chi}$ is a vector of variable selection weights. We remove the need for a static covariate encoder as we are not embedding static covariates within our model. For each time step, a non-linear processing is employed by feeding each ${\xi}_{t}^{(j)}$ through its own GRN:
\begin{align}
\tilde{{\xi}}_t^{(j)} = \text{GRN}_{\tilde{\xi}(j)}\left({\xi}_t^{(j)}\right), & \label{sec:varselect_grn}
\end{align}
where $\tilde{{\xi}}_t^{(j)}$ is the processed feature vector for variable $j$. We note that each variable has its own $\text{GRN}_{\xi(j)}$, with \emph{weights shared across all time steps $t$}. Processed features are then weighted by their variable selection weights and combined. This yields
\begin{align}
\tilde{{\xi}}_t  = \sum\nolimits_{j=1}^{m_\chi} v_{\chi_t}^{(j)}  \tilde{{\xi}}_t^{(j)}, &
\label{eq:var_selection_sum}
\end{align}
where $v_{\chi_t}^{(j)}$ is the j-th element of vector  ${v}_{\chi_t} $.

\subsubsection{Recurrent Block}

In a previous work \cite{genet2024tkan}, we introduced \emph{TKAN} to leverage the power of Kolmogorov-Arnold Network while offering memory management to handle time dependency. To introduce time dependency, each transformation function \(\phi_{l,j,i}\) has to be time dependent. Let us denote the sub layers memory state $\tilde{h}_{l,t}$, initialized with zeros and of shape $(\text{KAN}_{out})$.

\begin{figure}[H]
    \centering
    \includegraphics[width=1\linewidth]{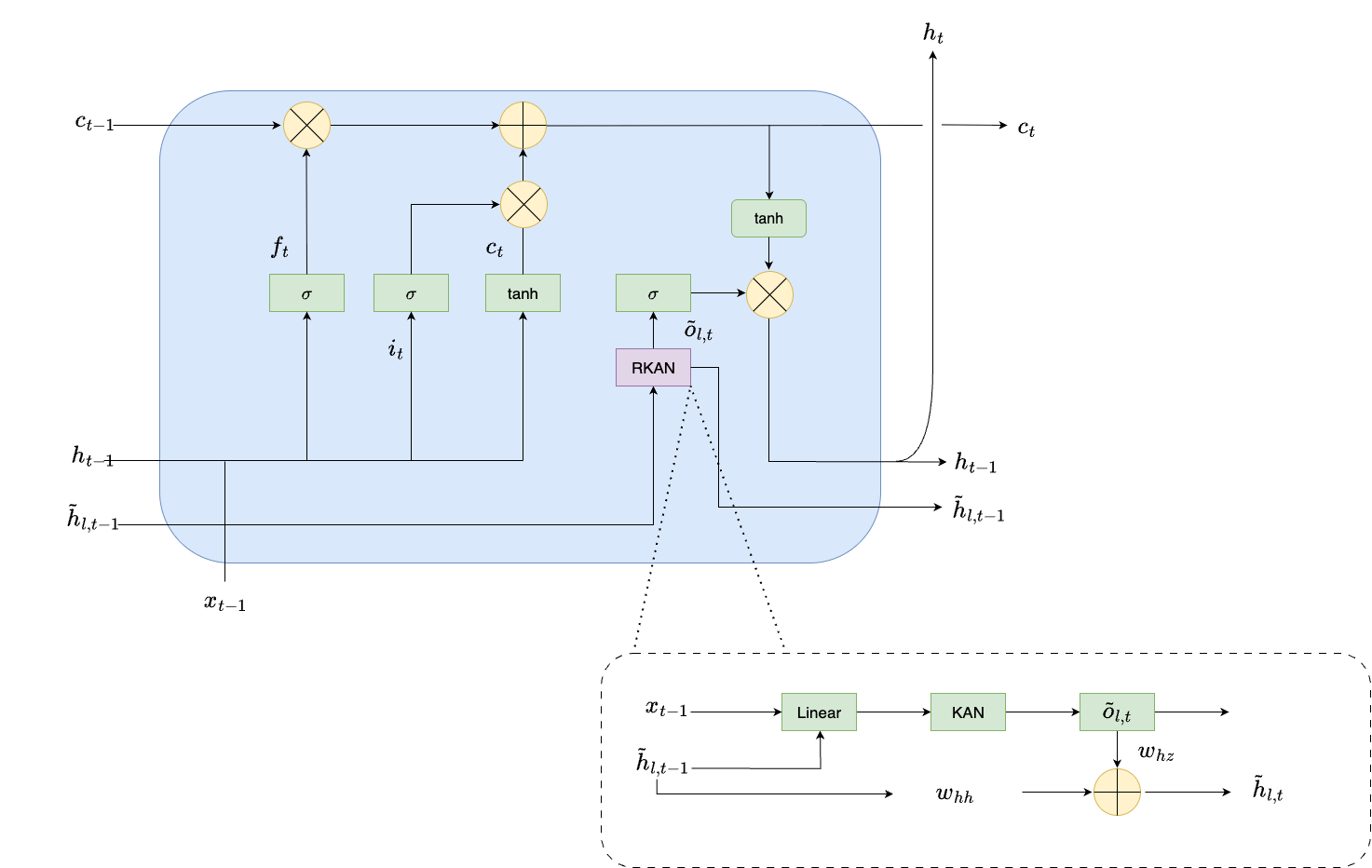}
    \caption{Temporal Kolmogorov-Arnold Networks (TKANs) \cite{genet2024tkan}} 
    \label{fig:tkan}
\end{figure}

The inputs that are fed to each sub KAN layers in the RKAN are created as follows:
\begin{equation}
    s_{l,t}=W_{l,\tilde{x}} x_t + W_{l,\tilde{h}} \tilde{h}_{l,t-1},
\end{equation}
where $W_{l,\tilde{x}}$ is the weight of the $l$-th layer applied to $x_t$, which is the input at time $t$. $W_{l,\tilde{h}}$ contain the weights of the $l$-th layer applied to its previous substate. Let us first denote $\text{KAN}_{in}$ and $\text{KAN}_{out}$ the input dimensions of RKAN Layer input and ouputs, respectively.
We have here $W_{l,\tilde{x}} \in \Rb^{(d, \text{KAN}_{in})} $ and $W_{l,\tilde{h}}^{(\text{KAN}_{out}, KAN_{in})}$, which leads to $s_{l,t}^{\text{KAN}_{in}}$.
\begin{equation}
\tilde{o}_{t} =  \phi_{l}(s_{l,t}),
\end{equation}
where $\phi_l$ is a KAN layer. The "memory" step \(\tilde{h}_{l,t}\) is defined as a combination of past hidden states, such as
\begin{equation}
\tilde{h}_{l,t} =  W_{hh} \tilde{h}_{l,t-1} + W_{hz} \tilde{o}_{t},
\label{eq:update_state_tkan}
\end{equation}
where $W$ is a vector of weights. $W$ weights the importance of past values relatively to the most recent inputs. For the next step, to maintain memory, we took inspiration from the LSTM \cite{hochreiter1997long,staudemeyer2019understanding} technique. We denote $x_t$ the input vector of dimension $d$. This unit uses several internal vectors and gates to manage information flow. The forget gate, with activation vector $f_t$, i.e., 
\begin{equation}
    f_t = \sigma(W_f x_t + U_f h_{t-1} + b_f),
\end{equation}
decides what information is forgotten from the previous state. The input gate, with activation vector denoted $i_t$, 
\begin{equation}
    i_t = \sigma(W_i x_t + U_i h_{t-1} + b_i),
\end{equation}
controls new information to be included. The output gate, with activation vector $o_t$,
\begin{equation}
    r_t = \text{Concat}[\phi_1(s_{1,t}),\phi_2(s_{2,t}),\ldots,\phi_L(s_{L,t})].
    \label{eq:concat_kan_out}
\end{equation}
$r_t$ concatenates the outputs of multiple KAN Layers and will fed an activation function to propagate a part of the information. Thus, we define
\begin{equation}
    o_t = \sigma(W_{o}r_t + b_o),
    \label{eq:out}
\end{equation}
where $W_{o} \in \Rb^{(\text{KAN}_out * L,out)}$, with $out$ the ouput dimension of TKAN. 
Equation \eqref{eq:out} determines the retained information from the current state to the output, given $r_t$ coming from \eqref{eq:concat_kan_out}. $\tilde{h}_t$ is the "sub" memory of the RKAN layers. The hidden state $h_t$ of the TKAN layer captures the unit's output, while the cell state $c_t$ is updated as
\begin{equation}
    c_t = f_t \odot c_{t-1} + i_t \odot \tilde{c}_t,
\end{equation}
where $\tilde{c}_t = \sigma(W_c x_t + U_c h_{t-1} + b_c)$
represents its internal memory. All these internal states have a dimensionality of $h$.
The ouput of the final hidden layer, denoted $h_t$ is given by
\begin{equation}
    h_t = o_t \odot \tanh(c_t).
    \label{eq:hidden_update}
\end{equation}

\subsubsection{Context \& Attention Block}

LEMs use a self-attention mechanism to capture long-term relationships between different time steps, modified from multi-head attention in transformer-based architectures \cite{li2019enhancing}. Critically, the attention mechanism employs a causal mask to prevent the model from accessing future information during context generation, ensuring that execution decisions are based only on historical and current market conditions.

\begin{align}
\text{Attention}(Q, K, V) &= A(Q, K) V, 
\label{attn}
\end{align}
where $A(.)$ is a normalization function. A common choice is the scaled dot-product attention ~\cite{vaswani2017attention}:
\begin{align}
A(Q, K) = \text{Softmax} \left( \frac{Q K^T}{\sqrt{d_{attn}}} \right).
\label{attn_normalize}
\end{align}
To enhance the learning capacity of the standard attention mechanism, \cite{vaswani2017attention} proposed multi-head attention, which employs different heads for different representation subspaces:
\begin{align}
\text{MultiHead} (Q, K, V) &= [H_1, \dots,  H_{m_H}] W_H, \\
H_h &= \text{Attention}(Q W_{Q}^{(h)},  K W_{K}^{(h)}, V W_{V}^{(h)}), 
\end{align}
where \(W_{K}^{(h)} \in \mathbb{R}^{d_{model} \times d_{attn}}\), \(W_{Q}^{(h)} \in \mathbb{R}^{d_{model} \times d_{attn}}\), and \(W_{V}^{(h)} \in \mathbb{R}^{d_{model} \times d_{V}}\) are head-specific weights for keys, queries, and values, respectively. The matrix \(W_H \in \mathbb{R}^{(m_H \cdot d_{V})\times d_{model}}\) linearly combines outputs concatenated from all heads \(H_h\). The causal masking ensures that attention weights $(A(Q,K))_{i,j} = 0$ for all $j > i$, preventing information leakage from future time steps during the context generation phase.

\subsection{Execution Decision Block}

\subsubsection{Step-wise FusedMLP Architecture}

The execution decision component represents one of the major innovations in this neural network architecture. Unlike traditional approaches that use a single decision module, LEMs employs a distinct FusedMLP at each execution step, acknowledging that optimal decisions vary depending on the temporal position within the execution timeline. Each step-specific FusedMLP receives a comprehensive input comprising the generated context from the Decision Context Generation Block, residual connections to the original input features to preserve critical information, all previous allocation decisions made in earlier steps, and real-time price and volume information at the current execution step.

Given an input denoted $\mathbf{X} \in \mathbb{R}^{B \times N \times F}$, where $B$ is the batch size, $N$ denotes the temporal horizon, and $F$ is the feature dimension, the model first replicates each feature vector across $M$ parallel paths, producing $X_{\text{expanded}} \in \mathbb{R}^{B \times N \times M \times F}$. Each of these $M$ fused paths corresponds to different execution modalities: volume versus notional allocation, VWAP versus TWAP benchmarking, and buy versus sell side decisions.

Each of these $M$ fused paths is processed independently through a succession of $L$ fully connected layers. At layer $\ell$, a distinct weight tensor $\mathbf{W}^{(\ell)} \in \mathbb{R}^{N \times M \times F_{\ell} \times H^{(\ell)}}$ and bias $\mathbf{b}^{(\ell)} \in \mathbb{R}^{N \times M \times H^{(\ell)}}$ are applied, yielding the transformation:
\begin{equation}
(\mathbf{X}^{(\ell)})_{b,n,m,h} = \phi^{(\ell)}\left( \sum_{f=1}^{F_{\ell}} (\mathbf{X}^{(\ell-1)})_{b,n,m,f} \cdot (\mathbf{W}^{(\ell)})_{n,m,f,h} + (\mathbf{b}^{(\ell)})_{n,m,h} \right),
\end{equation}
where $\phi^{(\ell)}$ is a configurable activation function. This formulation allows FusedMLP to simultaneously learn model-specific and modality-specific representations through parameterized, parallel transformations. The final output is a fused feature representation of shape $\mathbb{R}^{B \times N \times M \times H^{(L)}}$, enabling flexible integration of heterogeneous execution strategies while preserving distinctions across different allocation approaches.

\subsubsection{Constraint Handling and Differentiable Clipping}

To ensure realistic and feasible execution decisions, LEMs incorporates sophisticated constraint handling mechanisms. The model employs both hard and soft clipping techniques to maintain allocation constraints while preserving gradient flow during training. We define a differentiable soft clipping function $\mathcal{S}: \mathbb{R} \times \mathbb{R}^+ \times \mathbb{R}^+ \rightarrow \mathbb{R}$ that smoothly approaches constraint boundaries:
\begin{equation}
\mathcal{S}(x, u, \lambda) = x \cdot \sigma(\lambda \cdot (u - x)) + u \cdot (1 - \sigma(\lambda \cdot (u - x)))
\end{equation}
where $\sigma$ is the sigmoid function, $u$ represents the upper bound, and $\lambda > 0$ controls the transition sharpness. When $x \ll u$, the function approximates the identity $\mathcal{S}(x, u, \lambda) \approx x$, while when $x \gg u$, it approaches the upper bound $\mathcal{S}(x, u, \lambda) \approx u$.

At each execution step $s$, the model ensures that allocation decisions respect minimum and maximum trading rate constraints. Let $\alpha_s^{raw} \in \mathbb{R}^M$ denote the raw allocation output from the FusedMLP for step $s$, and let $r_{min}, r_{max} \in \mathbb{R}^M$ represent the minimum and maximum trading rates, respectively. The constrained allocation is computed as:
\begin{align}
\alpha_s^{temp} &= \mathcal{S}(\alpha_s^{raw} + r_{min}, r_{max}, \lambda) \\
\alpha_s &= \max(0, \min(\alpha_s^{temp}, R_s))
\end{align}
where $R_s$ represents the remaining budget at step $s$, and the $\max$ and $\min$ operations are applied element-wise.

The model maintains strict conservation of total allocation through a sequential allocation process. At each step, the remaining budget is updated as $R_{s+1} = R_s - \alpha_s$, with the constraint that $\sum_{s=1}^{S} \alpha_s = 1$, where $S$ is the total number of execution steps. The final step receives all remaining budget to ensure complete allocation: $\alpha_S = R_S$.

\subsubsection{Multi-Modal Output Generation}

The fundamental innovation of LEMs lies in its ability to train multiple order types and execution scenarios within a single unified framework. Rather than focusing on market impact minimization, the model addresses the benchmark beating problem across diverse execution conditions with time flexibility. The architecture simultaneously handles volume-based and notional-based execution strategies, VWAP and TWAP benchmarking approaches, and buy-side versus sell-side execution requirements. 

The key insight driving this multi-modal approach is that different execution scenarios fundamentally require the same market understanding and decision context. By sharing the decision context generation across all execution modalities, the model learns robust market representations that generalize across scenarios while avoiding overfitting to any specific execution type. Each execution scenario differs primarily in its allocation logic and benchmark comparison, but leverages the same underlying market pattern recognition capabilities.

Consider an order with a 12-period maximum horizon and a 6-period minimum execution requirement. The model learns to optimally allocate across these periods and potentially finish execution earlier than the maximum 12 steps to maximize gains against the benchmark. When execution terminates early, both the achieved price and benchmark computation end simultaneously at that point, allowing the model to capitalize on favorable market conditions by completing execution ahead of schedule while still meeting minimum duration constraints.

The final output tensor has shape $\mathbb{R}^{B \times T \times (T+1) \times 4 \times 2}$, where the dimensions represent batch size, execution steps, minimum periods, strategy combinations corresponding to Buy-VWAP, Buy-TWAP, Sell-VWAP, and Sell-TWAP execution modes, respectively and allocation type (volume or notional). Each combination represents a distinct execution scenario that requires beating its corresponding benchmark while respecting temporal constraints.

\subsection{Loss Function Formulation}

The loss function embodies the core objective of benchmark beating across multiple execution scenarios with time flexibility. Unlike traditional execution models that minimize market impact, LEMs maximizes profit-and-loss (PnL) by achieving execution prices superior to market benchmarks while managing associated risks.

Given predicted allocations $\mathbf{y}_{pred} \in \mathbb{R}^{B \times T \times (T+1) \times 4 \times 2}$ and market data $\mathbf{y}_{true} \in \mathbb{R}^{B \times T \times (T+1) \times 4 \times 2}$ where the last dimension contains market volumes and prices, the loss function operates through several key components.

\subsubsection{Execution Masking and Time Flexibility}

The model employs sophisticated masking to handle time flexibility constraints. A base triangular mask $\mathbf{M}_{base} \in \{0,1\}^{T \times T}$ ensures minimum execution duration requirements:
\begin{equation}
(\mathbf{M}_{base})_{i,j} = \begin{cases} 
1 & \text{if } j \leq i \\
0 & \text{otherwise}
\end{cases}
\end{equation}

This base mask is extended to include a final column of ones to represent the standard case where execution must complete, yielding $\mathbf{M} \in \{0,1\}^{B \times T \times (T+1) \times 4 \times 2}$. Technical masking prevents division by zero in benchmark computation when market volumes are zero:
\begin{equation}
(\mathbf{M}_{vol})_{b,t,n,m,k} = \begin{cases}
1 & \text{if } t = 1 \text{ or } \sum_{s=1}^{t-1} (V_{true})_{b,s,n,m} > 0 \\
0 & \text{otherwise}
\end{cases}
\end{equation}
where $V_{true}$ represents market volumes from $\mathbf{y}_{true}$.

\subsubsection{Soft Execution Decisions}

To enable gradient-based optimization, the model employs a soft masking mechanism that determines when execution is complete. For each execution path, we compute the reverse cumulative allocation to determine completion status:
\begin{equation}
(\mathbf{C})_{b,t,n,m,k} = \sum_{s=t}^{T} (\mathbf{y}_{pred})_{b,s,n,m,k}
\end{equation}
which represents the total remaining allocation from time $t$ onwards, indicating whether the order will be fully executed by time $t$. The soft completion indicator applies the sigmoid function to this cumulative measure:
\begin{equation}
\mathcal{M}_{soft}(\mathbf{C}, \alpha) = \sigma(\alpha \cdot \mathbf{C})
\end{equation}
where $\alpha$ controls the sharpness of the completion decision. The effective execution mask combines minimum time requirements with completion decisions:
\begin{equation}
\mathbf{M}_{eff} = \mathbf{M} + (1 - \mathbf{M}) \cdot \mathcal{M}_{soft}(\mathbf{C}, \alpha)
\end{equation}

\subsubsection{Price Achievement Calculation}

For volume target execution (first allocation type), executed volumes and achieved prices are:
\begin{align}
(\mathbf{V}_{exec})_{b,t,n,m} &= (\mathbf{y}_{pred})_{b,t,n,m,1} \odot (\mathbf{M}_{eff})_{b,t,n,m,1} \\
(\mathbf{N}_{exec})_{b,t,n,m} &= (\mathbf{V}_{exec})_{b,t,n,m} \odot (P_{market})_{b,t,n,m} \\
(\mathbf{P}_{achieved}^{vto})_{b,n,m} &= \frac{\sum_{t=1}^{T} (\mathbf{N}_{exec})_{b,t,n,m}}{\sum_{t=1}^{T} (\mathbf{V}_{exec})_{b,t,n,m} + \epsilon}
\end{align}
where $P_{market}$ represents market prices, $\odot$ denotes element-wise multiplication, and $\epsilon$ prevents division by zero.

For notional target execution (second allocation type):
\begin{align}
(\mathbf{N}_{exec})_{b,t,n,m} &= (\mathbf{y}_{pred})_{b,t,n,m,2} \odot (\mathbf{M}_{eff})_{b,t,n,m,2} \\
(\mathbf{V}_{exec})_{b,t,n,m} &= \frac{(\mathbf{N}_{exec})_{b,t,n,m}}{(P_{market})_{b,t,n,m}} \\
(\mathbf{P}_{achieved}^{nto})_{b,n,m} &= \frac{\sum_{t=1}^{T} (\mathbf{N}_{exec})_{b,t,n,m}}{\sum_{t=1}^{T} (\mathbf{V}_{exec})_{b,t,n,m} + \epsilon}
\end{align}

\subsubsection{Benchmark Computation}

Market benchmarks are computed over the effective execution periods, with technical adjustments to prevent division by zero:
\begin{align}
(\mathbf{V}_{market})_{b,t,n,m} &= (V_{true})_{b,t,n,m} \odot \max((\mathbf{M}_{eff})_{b,t,n,m}, (\mathbf{M}_{vol})_{b,t,n,m}) \\
(\mathbf{N}_{market})_{b,t,n,m} &= (\mathbf{V}_{market})_{b,t,n,m} \odot (P_{true})_{b,t,n,m} \\
(\mathbf{P}_{benchmark})_{b,n,m} &= \frac{\sum_{t=1}^{T} (\mathbf{N}_{market})_{b,t,n,m} + \epsilon}{\sum_{t=1}^{T} (\mathbf{V}_{market})_{b,t,n,m} + \epsilon}
\end{align}
where the volume mask $\mathbf{M}_{vol}$ ensures benchmark computation begins only after positive market volume, preventing numerical instabilities in the VWAP calculation.

\subsubsection{PnL Maximization Objective}

The core loss function maximizes PnL through asymmetric risk management using the softplus activation function. The softplus function is defined as:
\begin{equation}
\text{softplus}(x) = \ln(1 + e^x)
\end{equation}
This function exhibits asymmetric behavior that provides steeper penalties for losses while offering more moderate rewards for gains, effectively encouraging profit generation while discouraging excessive risk-taking.

The relative performance differences are computed as:
\begin{equation}
(\mathbf{D})_{b,a,n,m} = \left(\frac{(\mathbf{P}_{achieved})_{b,a,n,m}}{(\mathbf{P}_{benchmark})_{b,a,n,m}} - 1\right) \times 100
\end{equation}
where $a$ represents the allocation type (volume or notional).

The output tensor has $T+1$ minimum period dimensions, where the first $T$ dimensions correspond to flexible execution scenarios aimed at benchmark beating, and the final dimension represents a risk-only optimization scenario with no execution flexibility.

For the flexible execution scenarios (first $T$ minimum periods), the PnL component compares buy and sell execution performance:
\begin{equation}
\text{PnL} = \sum_{b=1}^{B} \sum_{a=1}^{2} \sum_{n=1}^{T} \sum_{k=1}^{2} \text{softplus}((\mathbf{D})_{b,a,n,k} - (\mathbf{D})_{b,a,n,k+2})
\end{equation}
where $a$ represents allocation type (volume or notional), indices $k \in \{1,2\}$ represent buy scenarios (Buy-VWAP, Buy-TWAP) and indices $k+2 \in \{3,4\}$ represent corresponding sell scenarios (Sell-VWAP, Sell-TWAP).

For the risk-only scenario (final minimum period), risk management is incorporated through:
\begin{equation}
\text{Risk} = \sum_{b=1}^{B} \sum_{a=1}^{2} \sum_{k=1}^{4} \text{softplus}(|(\mathbf{D})_{b,a,T+1,k}|)
\end{equation}

The final loss function combines both components:
\begin{equation}
\mathcal{L} = -\frac{\text{PnL} + \text{Risk}}{B \cdot 2 \cdot T \cdot 2 + B \cdot 2 \cdot 4}
\end{equation}
where the negative sign converts the maximization problem into a minimization objective suitable for gradient descent optimization, and the denominator normalizes by the total number of terms in each component.
\section{Learning task}
We evaluate our execution model on two distinct tasks: the first focuses on cryptocurrency markets with intraday execution horizons, while the second examines traditional equity markets from the Dow Jones Industrial Average over multi-day execution periods. On both tasks, the model is trained over the entire training set and, through its fused MLP architecture with separated logic paths, learns to handle all different order types simultaneously: VWAP or TWAP targets, volume or notional-based orders, as well as buy or sell orders, each with different minimum execution period constraints. We then analyze the model performance in each of these configurations.

\subsection{Cryptocurrency Market Dataset}

For the cryptocurrency evaluation, we utilize minutely candlestick data directly sourced from Binance Data Vision, encompassing twenty cryptocurrency spot trading pairs: ETHBTC, BNBBTC, BTCUSDT, ETHUSDT, LTCBTC, BNBUSDT, XRPBTC, LTCUSDT, XRPUSDT, ADAUSDT, ADABTC, LINKBTC, TRXUSDT, NEOUSDT, XLMUSDT, EOSUSDT, ETCUSDT, VETUSDT, ONTUSDT, and IOTAUSDT. From the candlestick data format, we extract only the volume and quote asset volume information, constructing our primary features from these two fundamental trading metrics.

\medskip

The deliberate inclusion of both USDT-denominated and Bitcoin-denominated pairs follows the multi-asset learning framework established in \cite{genet2025sigtransvwap}, enabling assessment of the model's capacity to learn unified execution patterns across different base currencies and market structures. This experimental design extends beyond traditional single-asset, single-frequency approaches by implementing simultaneous multi-asset and multi-frequency learning, generalizing the signature transformer concept by incorporating multiple temporal frequencies within a unified model architecture.

\medskip

The dataset encompasses data from August 1, 2017, through December 31, 2024, capturing approximately 7.4 years of continuous market activity across multiple cryptocurrency market cycles. Notably, not all trading pairs were available throughout the entire observation period, as assets were progressively listed on Binance during this timeframe, requiring careful handling of asset-specific data availability periods. We process this data at three distinct frequencies: 15-minute, 90-minute, and 250-minute intervals, enabling the model to capture execution patterns ranging from short-term microstructure effects to longer-term intraday trends.

\subsection{Equity Market Dataset}

For traditional equity markets, we focus on thirty stocks comprising the Dow Jones Industrial Average, specifically: AMZN, CVX, UNH, AMGN, VZ, AAPL, AXP, GS, SHW, JNJ, MCD, CSCO, IBM, WMT, PG, NKE, TRV, HON, MMM, MSFT, JPM, HD, KO, DIS, BA, CAT, NVDA, MRK, CRM, and V. The data spans from January 3, 2000, through December 31, 2024, sourced from Yahoo Finance. While quote asset volume information exists in equity markets, our data source does not provide direct access to these metrics. Consequently, we adopt a proxy approach by estimating quote asset volume through multiplication of trading volume with adjusted close prices. This approximation provides a reasonable substitute for notional trading activity while acknowledging the inherent limitations compared to exchange-native volume measurements.

\medskip

The temporal partitioning for equity data employs a validation split date of January 1, 2017, and a test split date of January 1, 2020. Unlike the cryptocurrency implementation, multiple frequencies are not merged in the equity evaluation, and we focus exclusively on 12 open trading day execution horizons, reflecting typical institutional execution timeframes in traditional markets.

\subsection{Feature Engineering}

Both cryptocurrency and equity datasets employ consistent feature engineering approaches following the volume adjustment methodology established in \cite{genet2025staticvwap, genet2024dynamic, genet2025sigtransvwap}. Volume normalization occurs through division by a rolling window of 336 data points shifted by the sum of lookback and prediction horizon lengths. This normalization strategy ensures scale-invariant learning across different market regimes and asset characteristics, though the specific time period represented varies with frequency.

\medskip

Price-based features center on returns computed from Volume Weighted Average Price (VWAP) values, providing a natural execution quality benchmark while capturing relative price movements independent of absolute price levels.Temporal feature engineering incorporates two seasonal components tailored to each market's characteristics. For cryptocurrency markets, we include hour-of-day and day-of-week features to capture well-documented intraday patterns and weekday effects in digital asset trading. For equity markets, we employ day-of-week and month-of-year features to capture traditional market seasonality, including phenomena such as reduced summer trading volumes and end-of-month portfolio rebalancing effects.

\medskip

These temporal features address established stylized facts in financial markets, such as the characteristic U-shaped intraday volume curves observed in cryptocurrency markets and seasonal trading patterns in traditional equity markets. The inclusion of these features enables the model to account for predictable cyclical behaviors while focusing learning capacity on more complex, non-seasonal execution dynamics.

\subsection{Training Configuration}

Our training approach imposes execution constraints without employing traditional data augmentation techniques. Specifically, we enforce a minimum trading requirement of $\frac{1}{n_{ahead}^2}$ per execution step when orders remain incomplete, while allowing maximum trading quantities to remain unconstrained. This constraint structure addresses practical risk management concerns, particularly the prevention of manipulation strategies where traders might delay execution until favorable market conditions materialize relative to benchmark prices.

\medskip

Training employs the Adam optimizer with conservative learning rates of 0.00001 for both cryptocurrency and equity datasets, reflecting the substantial dataset sizes and extended temporal coverage. For cryptocurrency markets, this configuration processes 958,976 training samples, 134,144 validation samples, and 133,888 test samples. The equity dataset comprises 100,864 training samples, 22,528 validation samples, and 37,376 test samples. These conservative learning rates ensure stable convergence across the diverse multi-asset training environments while preventing overfitting to specific market conditions or asset characteristics.

\medskip

We employ batches of size 256, and while the separation of training, validation, and test sets preserves temporal ordering in the data, we apply shuffling to the training set to ensure each batch contains a diverse mixture of data points across different time periods, frequencies, and assets. We implement early stopping and learning rate reduction on plateau with patience values of 2 and 1 epochs respectively, representing aggressive intervention thresholds. These values may appear conservative compared to typical deep learning applications, but were selected empirically after observing that the substantial dataset size leads to rapid overfitting when longer patience periods are employed, preventing models from being halted at their optimal performance points.

\subsection{Results on Dow Jones Stock}

We first present the results on Dow Jones stocks, as this task is simpler since it involves only one frequency. Here the goal is to dynamicaly allocate the order over 12 buisnes day, with a daily granularity.

\subsection{Tables}

Table \ref{tab:dow_comprehensive_performance} presents results for the model component that makes predictions for volume-denominated orders aiming to beat VWAP, showing results for both buying and selling across different minimum execution time constraints. For comparison, the first row shows equivalent results using a simple TWAP order denominated in either volume or notional.

\begin{table}[H]
\scriptsize
\caption{TWAP Benchmarks and VWAP Model Performance by Minimum Period}
\label{tab:dow_comprehensive_performance}
\begin{tabular}{l|l|rrrrrrr}
\toprule
 &  & Mean (bps) & Std (bps) & Median (bps) & P5 (bps) & P95 (bps) & Q25 (bps) & Q75 (bps) \\
Order Type & Min Period &  &  &  &  &  &  &  \\
\midrule
TWAP Notional & All & 0.00 & 59.60 & 0.78 & -68.38 & 64.78 & -13.14 & 15.26 \\
\cline{1-9}
TWAP Volume & All & 7.37 & 64.32 & 3.54 & -55.69 & 80.03 & -8.97 & 20.08 \\
\cline{1-9}
\multirow[t]{13}{*}{BUY} & 1 & -29.50 & 88.34 & -16.37 & -161.13 & 64.27 & -56.46 & 6.17 \\
 & 2 & -29.50 & 88.34 & -16.37 & -161.13 & 64.27 & -56.46 & 6.17 \\
 & 3 & -29.50 & 88.34 & -16.37 & -161.13 & 64.27 & -56.46 & 6.17 \\
 & 4 & -29.50 & 88.34 & -16.37 & -161.13 & 64.27 & -56.46 & 6.17 \\
 & 5 & -29.50 & 88.34 & -16.37 & -161.13 & 64.27 & -56.46 & 6.17 \\
 & 6 & -29.14 & 88.85 & -16.08 & -159.96 & 65.45 & -56.52 & 6.55 \\
 & 7 & -29.28 & 90.08 & -16.04 & -162.63 & 66.77 & -56.78 & 6.90 \\
 & 8 & -31.67 & 88.13 & -17.22 & -165.98 & 60.90 & -57.76 & 5.03 \\
 & 9 & -23.48 & 82.32 & -14.45 & -140.55 & 65.82 & -47.59 & 6.82 \\
 & 10 & -13.90 & 81.53 & -11.35 & -120.11 & 81.74 & -39.05 & 11.32 \\
 & 11 & -4.30 & 86.07 & -7.30 & -105.59 & 104.96 & -32.40 & 18.19 \\
 & 12 & 2.91 & 118.31 & -4.60 & -135.34 & 163.38 & -40.26 & 35.53 \\
 & Match VWAP & 9.61 & 53.46 & 6.26 & -47.18 & 76.26 & -8.88 & 24.88 \\
\cline{1-9}
\multirow[t]{13}{*}{SELL} & 1 & 36.33 & 85.67 & 23.56 & -39.38 & 152.76 & 3.90 & 53.36 \\
 & 2 & 36.33 & 85.67 & 23.56 & -39.38 & 152.76 & 3.90 & 53.36 \\
 & 3 & 36.33 & 85.67 & 23.56 & -39.38 & 152.76 & 3.90 & 53.36 \\
 & 4 & 36.33 & 85.67 & 23.56 & -39.38 & 152.76 & 3.90 & 53.36 \\
 & 5 & 36.33 & 85.67 & 23.56 & -39.38 & 152.76 & 3.90 & 53.36 \\
 & 6 & 36.23 & 85.50 & 23.52 & -39.45 & 151.70 & 3.83 & 53.43 \\
 & 7 & 35.95 & 86.47 & 23.40 & -40.87 & 153.52 & 3.63 & 53.45 \\
 & 8 & 37.55 & 86.77 & 24.40 & -38.21 & 155.23 & 4.43 & 54.60 \\
 & 9 & 32.57 & 76.90 & 22.59 & -40.53 & 134.32 & 3.72 & 49.51 \\
 & 10 & 25.64 & 70.90 & 18.82 & -47.82 & 118.35 & 0.11 & 43.47 \\
 & 11 & 17.40 & 68.13 & 13.32 & -59.27 & 104.02 & -5.45 & 36.26 \\
 & 12 & 6.58 & 88.71 & 7.18 & -101.50 & 112.99 & -18.52 & 31.87 \\
 & Match VWAP & 9.67 & 53.86 & 6.12 & -47.26 & 77.07 & -8.81 & 24.82 \\
\cline{1-9}
\bottomrule
\end{tabular}
\end{table}

This table demonstrates the model's ability to exploit execution time flexibility to outperform the benchmark. While TWAP orders used to execute against VWAP deviate slightly on average (+7bps) compared to the benchmark during the test period, the LEM achieves an average of -35bps for buying and +36bps for selling when given sufficient time flexibility (such as completing execution in half the allocated time). These results are achieved without substantially increased risk, as values at all percentile levels under this condition outperform the benchmark, representing a risk reduction despite a slightly higher standard deviation.

The table also show that the model is still not able to profit for too short minimum execution time, with results being sensibly similar for all 7 minimum end period values. However, the importance of the time freedom in the capacity of beating the benchmark appears while reducing it, indeed the average gains obtains reduce drastically when the model only have 2 steps of freedom, and in the case their is not time freedom the average gains become negligible in comparison to the increase in risk. Finally the last row represent the last dimension of the model that is fitted to match the VWAP, in this case it appears that the model is also able to reduce the execution risk in average, indeed compared to the TWAP in volume quantity the standard deviation is reduced by 17\%, with extreme values in 5\% and 95\% being also lowered. In the case where the order receive do not have time freedom, it is thus possibly more interesting to try to reduce the risk of it than to try to beat the benchmark in which case the risk become way more important.

\medskip

The table also shows that the model cannot generate more profits with very short minimum execution times, with results remaining similar across all minimum end period values from 0 to 4. However, the importance of temporal flexibility becomes apparent when examining its reduction: average gains decrease drastically when the model has only 2 steps of freedom, and when no time flexibility exists, average gains become negligible compared to the increased risk.

\medskip

The final row represents the model dimension fitted to match VWAP. Here, the model reduces execution risk on average: compared to volume-based TWAP, the standard deviation decreases by 17\%, with extreme values at the 5th and 95th percentiles also improving. When orders have no time flexibility, reducing execution risk may be more valuable than attempting to beat the benchmark, where risk becomes substantially higher.

\medskip

While the results under no flexibility constraints may appear disappointing, they actually indicate normal market behavior. The ability to consistently beat benchmarks on average without accepting additional risk would imply that a simple strategy of simultaneously buying and selling identical orders could generate risk-free profits. Such an outcome would be highly improbable in practice unless the model were generating genuine alpha, which is not the case here.

\medskip

The following tables present results for each minimum execution end period across different model objectives:

\begin{table}[H]
\scriptsize
\caption{Detailed Model Comparison (Minimum Period 7)}
\label{tab:dow_detailed_min_period_6}
\begin{tabular}{l|l|rrrrrrr}
\toprule
 &  & Mean (bps) & Std (bps) & Median (bps) & P5 (bps) & P95 (bps) & Q25 (bps) & Q75 (bps) \\
Order Type & Strategy &  &  &  &  &  &  &  \\
\midrule
\multirow[t]{4}{*}{BUY} & TWAP Volume vs TWAP & -37.42 & 59.53 & -19.15 & -147.80 & 19.17 & -61.46 & -0.51 \\
 & TWAP Notional vs TWAP & -37.90 & 59.06 & -22.15 & -145.08 & 23.20 & -61.61 & -1.19 \\
 & VWAP Volume vs VWAP & -29.28 & 90.08 & -16.04 & -162.63 & 66.77 & -56.78 & 6.90 \\
 & VWAP Notional vs VWAP & -30.78 & 90.75 & -16.94 & -167.95 & 66.97 & -56.69 & 6.23 \\
\cline{1-9}
\multirow[t]{4}{*}{SELL} & TWAP Volume vs TWAP & 30.33 & 47.50 & 21.22 & -16.39 & 109.26 & 6.23 & 43.87 \\
 & TWAP Notional vs TWAP & 24.07 & 41.95 & 19.17 & -24.45 & 91.84 & 5.06 & 38.99 \\
 & VWAP Volume vs VWAP & 35.95 & 86.47 & 23.40 & -40.87 & 153.52 & 3.63 & 53.45 \\
 & VWAP Notional vs VWAP & 30.48 & 75.55 & 21.85 & -46.70 & 137.79 & 2.85 & 48.89 \\
\cline{1-9}
\bottomrule
\end{tabular}
\end{table}

Table \ref{tab:dow_detailed_min_period_6} shows consistent model performance across execution types. Notably, even for TWAP targets, the model learns and performs out-of-sample with relative gain magnitudes similar to VWAP-targeted orders. Since TWAP is deterministic, the risk of perfect matching should theoretically be zero (though order splitting and execution slippage not directly represented here would create deviations in practice). The non-zero risk therefore indicates that the model accepts additional risk to outperform its benchmark.

\begin{table}[H]
\scriptsize
\caption{Detailed Model Comparison (Minimum Period 10)}
\label{tab:dow_detailed_min_period_9}
\begin{tabular}{l|l|rrrrrrr}
\toprule
 &  & Mean (bps) & Std (bps) & Median (bps) & P5 (bps) & P95 (bps) & Q25 (bps) & Q75 (bps) \\
Order Type & Strategy &  &  &  &  &  &  &  \\
\midrule
\multirow[t]{4}{*}{BUY} & TWAP Volume vs TWAP & -20.39 & 40.13 & -16.06 & -87.53 & 32.93 & -39.27 & 1.60 \\
 & TWAP Notional vs TWAP & -26.64 & 41.40 & -19.30 & -96.17 & 19.50 & -41.31 & -3.80 \\
 & VWAP Volume vs VWAP & -13.90 & 81.53 & -11.35 & -120.11 & 81.74 & -39.05 & 11.32 \\
 & VWAP Notional vs VWAP & -20.16 & 74.64 & -13.43 & -124.52 & 62.88 & -40.46 & 6.73 \\
\cline{1-9}
\multirow[t]{4}{*}{SELL} & TWAP Volume vs TWAP & 18.51 & 36.40 & 16.10 & -30.77 & 75.69 & 2.41 & 32.71 \\
 & TWAP Notional vs TWAP & 12.13 & 42.25 & 14.31 & -48.45 & 63.80 & -0.30 & 29.87 \\
 & VWAP Volume vs VWAP & 25.64 & 70.90 & 18.82 & -47.82 & 118.35 & 0.11 & 43.47 \\
 & VWAP Notional vs VWAP & 19.17 & 62.28 & 16.95 & -59.37 & 102.12 & -2.07 & 39.73 \\
\cline{1-9}
\bottomrule
\end{tabular}
\end{table}

\begin{table}[H]
\scriptsize
\caption{Detailed Model Comparison (Minimum Period 12)}
\label{tab:dow_detailed_min_period_11}
\begin{tabular}{l|l|rrrrrrr}
\toprule
 &  & Mean (bps) & Std (bps) & Median (bps) & P5 (bps) & P95 (bps) & Q25 (bps) & Q75 (bps) \\
Order Type & Strategy &  &  &  &  &  &  &  \\
\midrule
\multirow[t]{4}{*}{BUY} & TWAP Volume vs TWAP & -4.75 & 69.65 & -8.92 & -103.64 & 104.44 & -39.30 & 26.03 \\
 & TWAP Notional vs TWAP & -0.05 & 123.71 & 1.84 & -168.24 & 166.38 & -46.60 & 53.67 \\
 & VWAP Volume vs VWAP & 2.91 & 118.31 & -4.60 & -135.34 & 163.38 & -40.26 & 35.53 \\
 & VWAP Notional vs VWAP & 4.90 & 88.58 & 5.12 & -122.07 & 134.03 & -38.25 & 51.48 \\
\cline{1-9}
\multirow[t]{4}{*}{SELL} & TWAP Volume vs TWAP & -0.91 & 45.48 & 3.18 & -75.55 & 59.86 & -18.36 & 20.54 \\
 & TWAP Notional vs TWAP & 4.94 & 160.23 & 22.02 & -242.56 & 212.90 & -58.63 & 83.51 \\
 & VWAP Volume vs VWAP & 6.58 & 88.71 & 7.18 & -101.50 & 112.99 & -18.52 & 31.87 \\
 & VWAP Notional vs VWAP & 9.75 & 120.21 & 20.68 & -182.05 & 174.48 & -47.97 & 77.45 \\
\cline{1-9}
\bottomrule
\end{tabular}
\end{table}

Tables \ref{tab:dow_detailed_min_period_9} and \ref{tab:dow_detailed_min_period_11} confirm that the patterns observed in the main table hold across model types, with cases lacking time flexibility resulting in disproportionately high risk relative to average gains.

\begin{table}[H]
\scriptsize
\caption{Detailed Model Comparison (Match Benchmark)}
\label{tab:dow_detailed_min_period_12}
\begin{tabular}{l|l|rrrrrrr}
\toprule
 &  & Mean (bps) & Std (bps) & Median (bps) & P5 (bps) & P95 (bps) & Q25 (bps) & Q75 (bps) \\
Order Type & Strategy &  &  &  &  &  &  &  \\
\midrule
\multirow[t]{4}{*}{BUY} & TWAP Volume vs TWAP & 0.04 & 4.10 & 0.04 & -4.27 & 4.21 & -0.96 & 1.03 \\
 & TWAP Notional vs TWAP & -6.96 & 23.08 & -2.37 & -28.59 & 2.43 & -7.20 & -0.25 \\
 & VWAP Volume vs VWAP & 9.61 & 53.46 & 6.26 & -47.18 & 76.26 & -8.88 & 24.88 \\
 & VWAP Notional vs VWAP & 1.88 & 50.44 & 2.83 & -58.73 & 61.08 & -12.83 & 19.09 \\
\cline{1-9}
\multirow[t]{4}{*}{SELL} & TWAP Volume vs TWAP & -0.03 & 4.55 & 0.01 & -4.32 & 4.10 & -0.98 & 0.99 \\
 & TWAP Notional vs TWAP & -7.18 & 23.29 & -2.52 & -30.02 & 2.18 & -7.37 & -0.35 \\
 & VWAP Volume vs VWAP & 9.67 & 53.86 & 6.12 & -47.26 & 77.07 & -8.81 & 24.82 \\
 & VWAP Notional vs VWAP & 1.83 & 50.59 & 2.74 & -58.62 & 61.11 & -12.82 & 18.97 \\
\cline{1-9}
\bottomrule
\end{tabular}
\end{table}

Finally, Table \ref{tab:dow_detailed_min_period_12}, where the goal is matching the benchmark, shows that the model reduces VWAP execution risk compared to simple TWAP strategies. However, the non-zero risk in TWAP Volume vs TWAP highlights imperfect fitting, as optimal results should theoretically be zero. Using deep learning to predict a constant target represents unnecessary complexity, and these results serve primarily to demonstrate suboptimal model fitting rather than practical utility.

\subsection{Distributions}

\begin{figure}[H]
    \centering
 \includegraphics[scale=0.4]{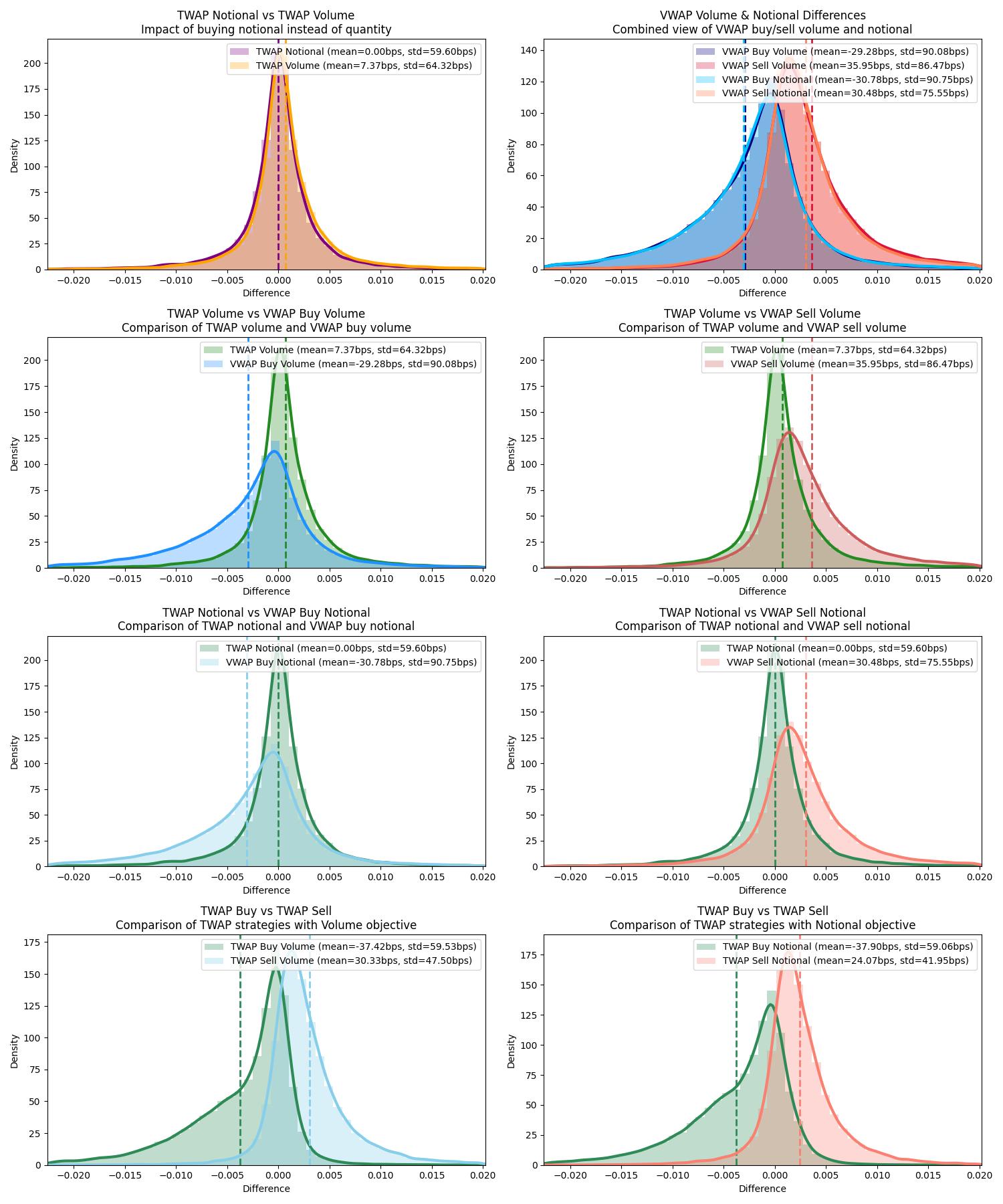}
  \caption{Dow Jones - Slippage Distribution - Min 7 periods}
    \label{fig:dow_6}
   
\end{figure}

Examining the distribution of benchmark slippage across the test set provides additional insight into model performance. Figure \ref{fig:dow_6} illustrates how execution end flexibility creates highly skewed distributions for buying and selling. The figure also demonstrates the well-known convexity effect of notional versus volume orders, visible in the slight deviation between the two TWAP curves.

\begin{figure}[H]
    \centering
 \includegraphics[scale=0.4]{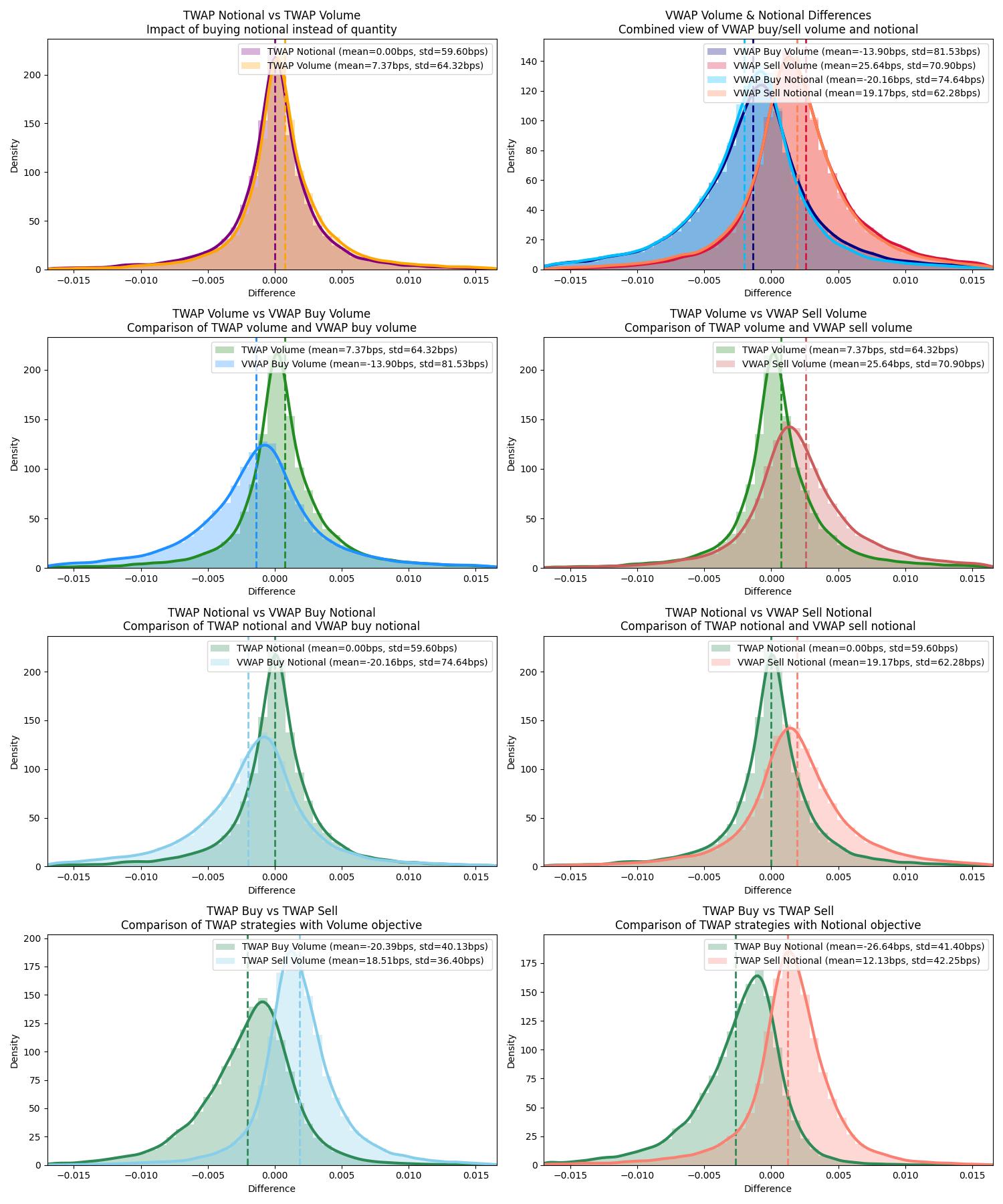}
 \caption{Dow Jones - Slippage Distribution - Min 10 periods}
    \label{fig:dow_9}
    
\end{figure}

\begin{figure}[H]
    \centering
 \includegraphics[scale=0.4]{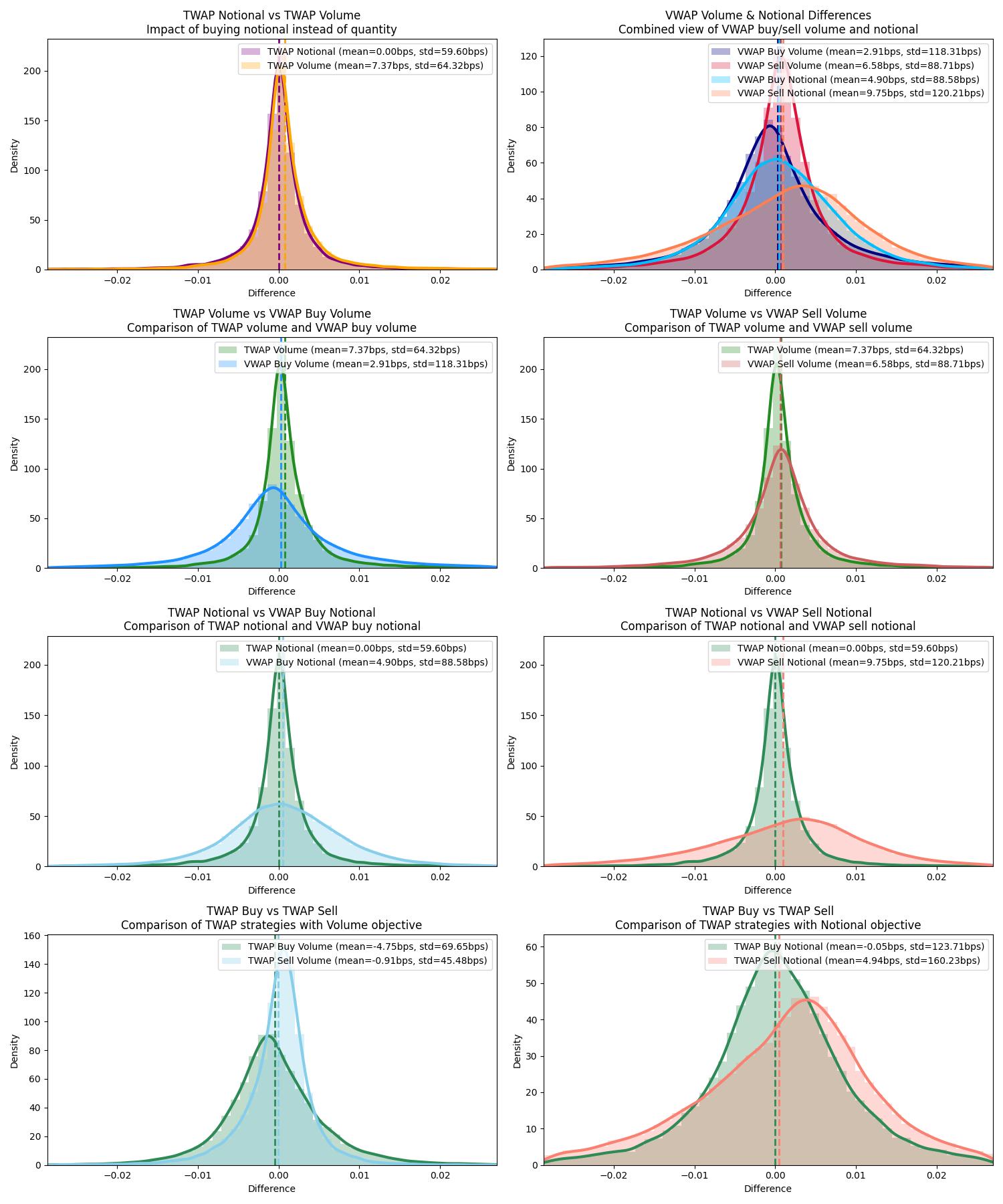}
 \caption{Dow Jones - Slippage Distribution - Min 12 periods}
    \label{fig:dow_11}
    
\end{figure}

Figures \ref{fig:dow_9} and \ref{fig:dow_11} demonstrate the patterns observed in the preceding tables. As flexibility decreases, the curves converge and distributions become less skewed. In the extreme case without flexibility, distributions become substantially wider, highlighting the model's inability to contain risk while beating benchmarks and discouraging its use under such constraints.

\begin{figure}[H]
    \centering
 \includegraphics[scale=0.4]{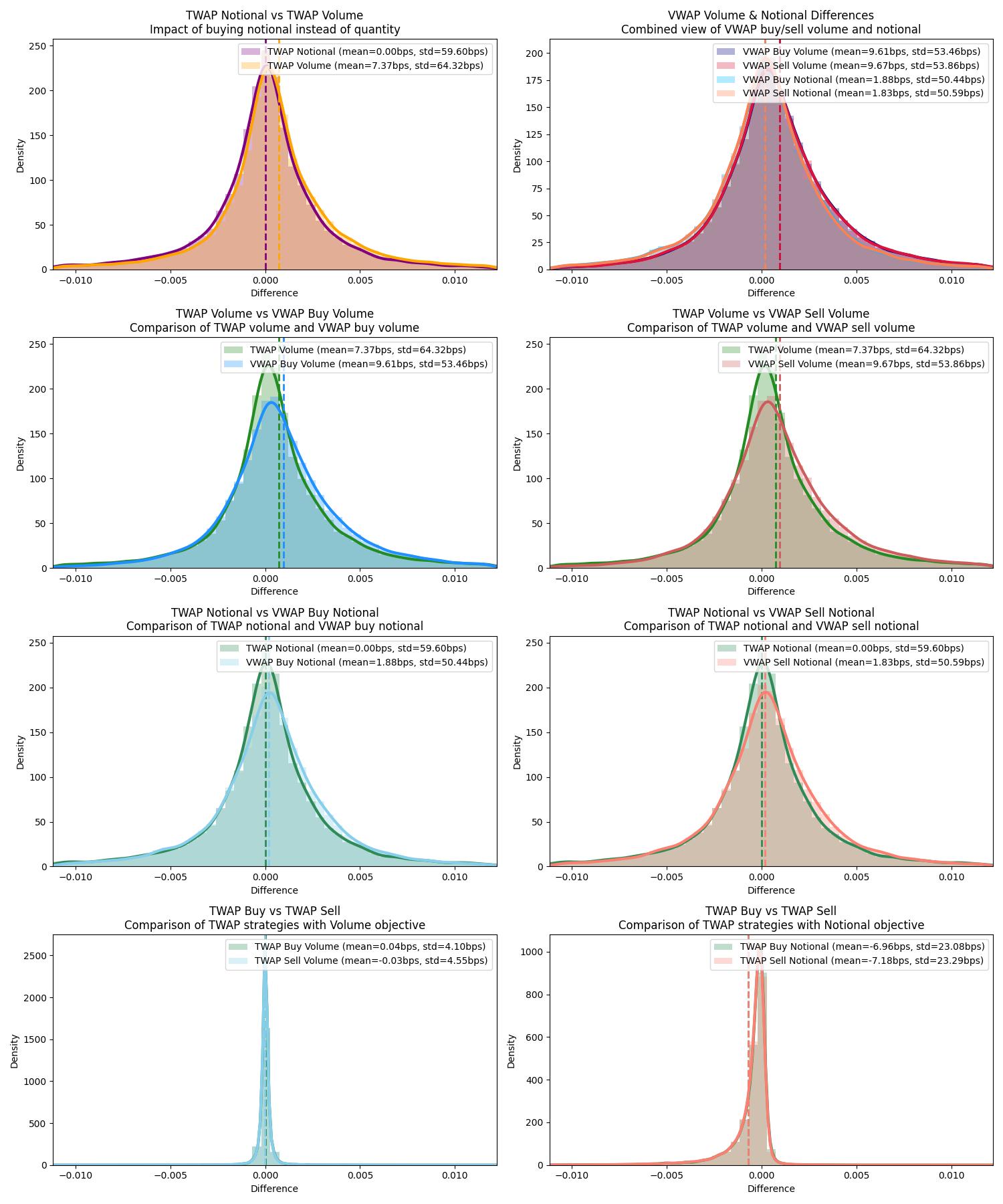}
  \caption{Dow Jones - Slippage Distribution - Match VWAP}
    \label{fig:dow_12}
   
\end{figure}

Figure \ref{fig:dow_12} shows how the VWAP-matching component offers superior risk reduction compared to the VWAP-beating component. While the TWAP order slippage curves appear more peaked and seemingly more centered, examination of the previous tables (particularly extreme cases beyond these boundaries) provides context. The VWAP-matching model accepts higher risk in low-risk scenarios to hedge against more severe deviations. This behavior matches patterns observed in previous papers, particularly the end-loaded curves explained in \cite{genet2025staticvwap, genet2024dynamic}.

\subsection{Results on digital assets}

This second task presents an even more interesting challenge, as it involves training a unified model across different frequencies: 15, 90, and 250 minutes, respectively, with twelve allocation decisions to make at each frequency. The objective is to test whether it is possible to train a single model that covers a wide range of cases simultaneously.

\subsection{Tables}

Table \ref{tab:crypto_15_comprehensive_performance} presents the results for the smallest frequency of 15 minutes. The table reveals several interesting patterns similar to those observed for stocks, where the lowest time duration constraint presents similar values across the first seven minimum periods, with an average buy slippage 7 basis points lower than the benchmark and average sell price nearly 9 basis points above the benchmark. The pattern of declining average gains as flexibility reduces is also evident. However, what is particularly noteworthy is that the standard deviation also diminishes when flexibility is reduced, indicating that the proposed loss function, which incorporates both performance and risk components, performs better on cryptocurrency data. It is even observable that the model without time freedom is able to achieve better slippage on both buy and sell orders compared to the TWAP benchmark, while simultaneously reducing execution risk. Finally, we observe that the Match VWAP component outperforms all others in terms of risk, with a 20\% reduction in risk.

\begin{table}[H]
\scriptsize
\caption{TWAP Benchmarks and VWAP Model Performance by Minimum Period with frequency: 15 minutes}
\label{tab:crypto_15_comprehensive_performance}
\begin{tabular}{l|l|rrrrrrr}
\toprule
 &  & Mean (bps) & Std (bps) & Median (bps) & P5 (bps) & P95 (bps) & Q25 (bps) & Q75 (bps) \\
Order Type & Min Period &  &  &  &  &  &  &  \\
\midrule
TWAP Notional & All & 0.93 & 30.54 & 0.39 & -26.55 & 29.70 & -4.44 & 6.09 \\
\cline{1-9}
TWAP Volume & All & 1.30 & 30.52 & 0.46 & -25.84 & 30.42 & -4.31 & 6.24 \\
\cline{1-9}
\multirow[t]{13}{*}{BUY} & 1 & -6.96 & 32.64 & -2.53 & -43.15 & 17.68 & -10.27 & 2.07 \\
 & 2 & -6.96 & 32.64 & -2.53 & -43.15 & 17.68 & -10.27 & 2.07 \\
 & 3 & -6.96 & 32.64 & -2.53 & -43.15 & 17.68 & -10.27 & 2.07 \\
 & 4 & -6.96 & 32.64 & -2.53 & -43.15 & 17.68 & -10.27 & 2.07 \\
 & 5 & -6.96 & 32.64 & -2.53 & -43.19 & 17.69 & -10.26 & 2.07 \\
 & 6 & -6.95 & 32.66 & -2.54 & -43.12 & 17.68 & -10.25 & 2.06 \\
 & 7 & -6.96 & 32.66 & -2.53 & -43.16 & 17.61 & -10.27 & 2.06 \\
 & 8 & -6.80 & 32.55 & -2.47 & -42.75 & 17.77 & -10.17 & 2.11 \\
 & 9 & -6.09 & 31.19 & -2.14 & -40.73 & 18.38 & -9.37 & 2.36 \\
 & 10 & -4.38 & 29.67 & -1.56 & -36.09 & 20.14 & -8.21 & 3.11 \\
 & 11 & -2.53 & 27.84 & -0.94 & -31.08 & 21.77 & -7.12 & 4.07 \\
 & 12 & -1.16 & 25.99 & -0.21 & -31.20 & 26.60 & -7.81 & 6.56 \\
 & Match VWAP & 1.13 & 23.97 & 0.56 & -22.33 & 25.99 & -4.63 & 6.14 \\
\cline{1-9}
\multirow[t]{13}{*}{SELL} & 1 & 8.80 & 35.00 & 3.51 & -16.92 & 49.39 & -1.59 & 12.87 \\
 & 2 & 8.80 & 35.00 & 3.51 & -16.92 & 49.39 & -1.59 & 12.87 \\
 & 3 & 8.80 & 35.00 & 3.51 & -16.92 & 49.39 & -1.59 & 12.87 \\
 & 4 & 8.80 & 35.00 & 3.51 & -16.92 & 49.39 & -1.59 & 12.87 \\
 & 5 & 8.80 & 34.99 & 3.51 & -16.91 & 49.40 & -1.59 & 12.87 \\
 & 6 & 8.81 & 35.03 & 3.52 & -16.87 & 49.45 & -1.59 & 12.87 \\
 & 7 & 8.81 & 35.05 & 3.51 & -16.82 & 49.56 & -1.60 & 12.86 \\
 & 8 & 8.63 & 34.80 & 3.43 & -17.04 & 48.92 & -1.67 & 12.68 \\
 & 9 & 7.95 & 34.12 & 3.13 & -17.37 & 46.71 & -1.80 & 11.98 \\
 & 10 & 6.51 & 31.79 & 2.62 & -18.02 & 42.23 & -2.11 & 10.60 \\
 & 11 & 4.93 & 29.22 & 1.97 & -19.09 & 37.00 & -2.70 & 9.28 \\
 & 12 & 3.78 & 27.48 & 1.29 & -22.26 & 36.24 & -4.48 & 9.18 \\
 & Match VWAP & 1.12 & 23.95 & 0.56 & -22.37 & 25.99 & -4.63 & 6.15 \\
\cline{1-9}
\bottomrule
\end{tabular}
\end{table}

Tables \ref{tab:crypto_90_comprehensive_performance} and \ref{tab:crypto_250_comprehensive_performance} present the results of the same model on inputs of different frequencies and confirm the patterns observed in the previous tables. Both risk and performance values scale with the square root of the time scaling increase, which is entirely logical as variance also increases by the square root of time. Notably, both average order-to-benchmark deviation and standard deviation of this deviation scale proportionally, meaning that the performance-to-risk ratio remains consistent. Finally, all value comparisons to the TWAP benchmark order remain consistent, indicating that it is indeed possible to merge multiple frequencies into one unified model.

\begin{table}[H]
\scriptsize
\caption{TWAP Benchmarks and VWAP Model Performance by Minimum Period with frequency: 90 minutes}
\label{tab:crypto_90_comprehensive_performance}
\begin{tabular}{l|l|rrrrrrr}
\toprule
 &  & Mean (bps) & Std (bps) & Median (bps) & P5 (bps) & P95 (bps) & Q25 (bps) & Q75 (bps) \\
Order Type & Min Period &  &  &  &  &  &  &  \\
\midrule
TWAP Notional & All & -1.06 & 65.07 & 0.16 & -72.06 & 63.24 & -12.34 & 12.24 \\
\cline{1-9}
TWAP Volume & All & 1.05 & 64.01 & 0.56 & -67.78 & 66.61 & -11.48 & 13.08 \\
\cline{1-9}
\multirow[t]{13}{*}{BUY} & 1 & -18.57 & 76.72 & -6.93 & -112.03 & 39.98 & -26.71 & 3.77 \\
 & 2 & -18.57 & 76.72 & -6.93 & -112.03 & 39.98 & -26.71 & 3.77 \\
 & 3 & -18.57 & 76.72 & -6.93 & -112.03 & 39.98 & -26.71 & 3.77 \\
 & 4 & -18.57 & 76.72 & -6.93 & -112.03 & 39.98 & -26.71 & 3.77 \\
 & 5 & -18.56 & 76.71 & -6.92 & -112.02 & 40.01 & -26.71 & 3.78 \\
 & 6 & -18.54 & 76.65 & -6.95 & -111.83 & 40.10 & -26.68 & 3.77 \\
 & 7 & -18.54 & 76.77 & -6.90 & -112.13 & 40.17 & -26.74 & 3.79 \\
 & 8 & -18.18 & 76.23 & -6.83 & -110.75 & 40.60 & -26.53 & 3.90 \\
 & 9 & -16.04 & 73.25 & -5.79 & -103.78 & 40.90 & -24.20 & 4.56 \\
 & 10 & -12.32 & 67.82 & -4.67 & -92.12 & 43.52 & -21.19 & 5.92 \\
 & 11 & -8.11 & 60.88 & -3.41 & -77.93 & 46.65 & -18.51 & 7.72 \\
 & 12 & -4.65 & 56.16 & -2.10 & -73.60 & 55.15 & -20.33 & 13.29 \\
 & Match VWAP & 0.90 & 50.48 & 0.30 & -56.59 & 56.73 & -11.88 & 12.48 \\
\cline{1-9}
\multirow[t]{13}{*}{SELL} & 1 & 20.43 & 73.93 & 8.53 & -40.43 & 115.36 & -3.53 & 30.54 \\
 & 2 & 20.43 & 73.93 & 8.53 & -40.43 & 115.36 & -3.53 & 30.54 \\
 & 3 & 20.43 & 73.93 & 8.53 & -40.43 & 115.36 & -3.53 & 30.54 \\
 & 4 & 20.43 & 73.93 & 8.53 & -40.43 & 115.36 & -3.53 & 30.54 \\
 & 5 & 20.42 & 73.92 & 8.53 & -40.40 & 115.35 & -3.52 & 30.54 \\
 & 6 & 20.42 & 73.93 & 8.54 & -40.52 & 115.36 & -3.53 & 30.50 \\
 & 7 & 20.42 & 74.06 & 8.50 & -40.65 & 115.58 & -3.57 & 30.58 \\
 & 8 & 19.87 & 73.47 & 8.35 & -41.33 & 113.79 & -3.65 & 30.10 \\
 & 9 & 17.77 & 71.60 & 7.22 & -43.80 & 108.33 & -4.46 & 27.72 \\
 & 10 & 14.14 & 66.44 & 5.92 & -45.68 & 94.99 & -5.12 & 23.98 \\
 & 11 & 10.02 & 59.25 & 4.14 & -47.53 & 80.99 & -6.82 & 19.92 \\
 & 12 & 7.08 & 55.43 & 2.43 & -53.92 & 78.57 & -11.31 & 19.74 \\
 & Match VWAP & 0.90 & 50.42 & 0.30 & -56.53 & 56.92 & -11.91 & 12.51 \\
\cline{1-9}
\bottomrule
\end{tabular}
\end{table}

\begin{table}[H]
\scriptsize
\caption{TWAP Benchmarks and VWAP Model Performance by Minimum Period with frequency: 250 minutes}
\label{tab:crypto_250_comprehensive_performance}
\begin{tabular}{l|l|rrrrrrr}
\toprule
 &  & Mean (bps) & Std (bps) & Median (bps) & P5 (bps) & P95 (bps) & Q25 (bps) & Q75 (bps) \\
Order Type & Min Period &  &  &  &  &  &  &  \\
\midrule
TWAP Notional & All & -7.04 & 109.62 & -1.94 & -124.01 & 97.34 & -23.17 & 16.86 \\
\cline{1-9}
TWAP Volume & All & -1.15 & 103.97 & -0.75 & -111.59 & 106.71 & -20.74 & 19.02 \\
\cline{1-9}
\multirow[t]{13}{*}{BUY} & 1 & -35.52 & 130.46 & -14.68 & -183.08 & 57.83 & -46.54 & 3.66 \\
 & 2 & -35.52 & 130.46 & -14.68 & -183.08 & 57.83 & -46.54 & 3.66 \\
 & 3 & -35.52 & 130.46 & -14.68 & -183.08 & 57.83 & -46.54 & 3.66 \\
 & 4 & -35.52 & 130.46 & -14.68 & -183.08 & 57.83 & -46.54 & 3.66 \\
 & 5 & -35.48 & 130.50 & -14.69 & -183.05 & 58.05 & -46.53 & 3.69 \\
 & 6 & -35.30 & 130.31 & -14.67 & -182.79 & 58.35 & -46.46 & 3.72 \\
 & 7 & -35.28 & 130.26 & -14.71 & -183.19 & 58.78 & -46.43 & 3.69 \\
 & 8 & -34.53 & 130.03 & -14.57 & -180.17 & 58.45 & -45.47 & 3.90 \\
 & 9 & -30.44 & 125.18 & -12.36 & -170.93 & 61.56 & -41.26 & 4.83 \\
 & 10 & -23.80 & 113.85 & -10.29 & -147.35 & 62.43 & -36.39 & 7.27 \\
 & 11 & -16.23 & 100.05 & -7.81 & -128.66 & 71.02 & -33.43 & 11.69 \\
 & 12 & -10.31 & 92.45 & -5.13 & -128.74 & 90.55 & -38.30 & 22.38 \\
 & Match VWAP & -0.61 & 82.33 & -1.20 & -94.88 & 92.94 & -21.34 & 18.27 \\
\cline{1-9}
\multirow[t]{13}{*}{SELL} & 1 & 33.60 & 116.11 & 14.74 & -65.09 & 193.33 & -5.68 & 49.40 \\
 & 2 & 33.60 & 116.11 & 14.74 & -65.09 & 193.33 & -5.68 & 49.40 \\
 & 3 & 33.60 & 116.11 & 14.74 & -65.09 & 193.33 & -5.68 & 49.40 \\
 & 4 & 33.60 & 116.11 & 14.74 & -65.09 & 193.33 & -5.68 & 49.40 \\
 & 5 & 33.61 & 116.16 & 14.74 & -65.09 & 193.78 & -5.68 & 49.42 \\
 & 6 & 33.68 & 116.09 & 14.77 & -65.18 & 193.85 & -5.61 & 49.42 \\
 & 7 & 33.75 & 116.40 & 14.68 & -65.06 & 194.47 & -5.72 & 49.62 \\
 & 8 & 32.50 & 114.98 & 14.34 & -66.33 & 191.65 & -6.04 & 48.82 \\
 & 9 & 27.96 & 113.01 & 12.08 & -70.04 & 180.29 & -7.65 & 44.08 \\
 & 10 & 21.23 & 102.39 & 9.54 & -71.41 & 154.18 & -9.15 & 37.44 \\
 & 11 & 13.88 & 90.83 & 6.24 & -77.06 & 130.81 & -12.87 & 31.46 \\
 & 12 & 9.06 & 87.43 & 2.44 & -93.20 & 132.60 & -22.33 & 33.04 \\
 & Match VWAP & -0.57 & 82.19 & -1.22 & -94.70 & 93.01 & -21.21 & 18.23 \\
\cline{1-9}
\bottomrule
\end{tabular}
\end{table}

Additional tables similar to those in the stock section are available on the paper github repository.

\subsection{Distributions}

\begin{figure}[H]
    \centering
 \includegraphics[scale=0.4]{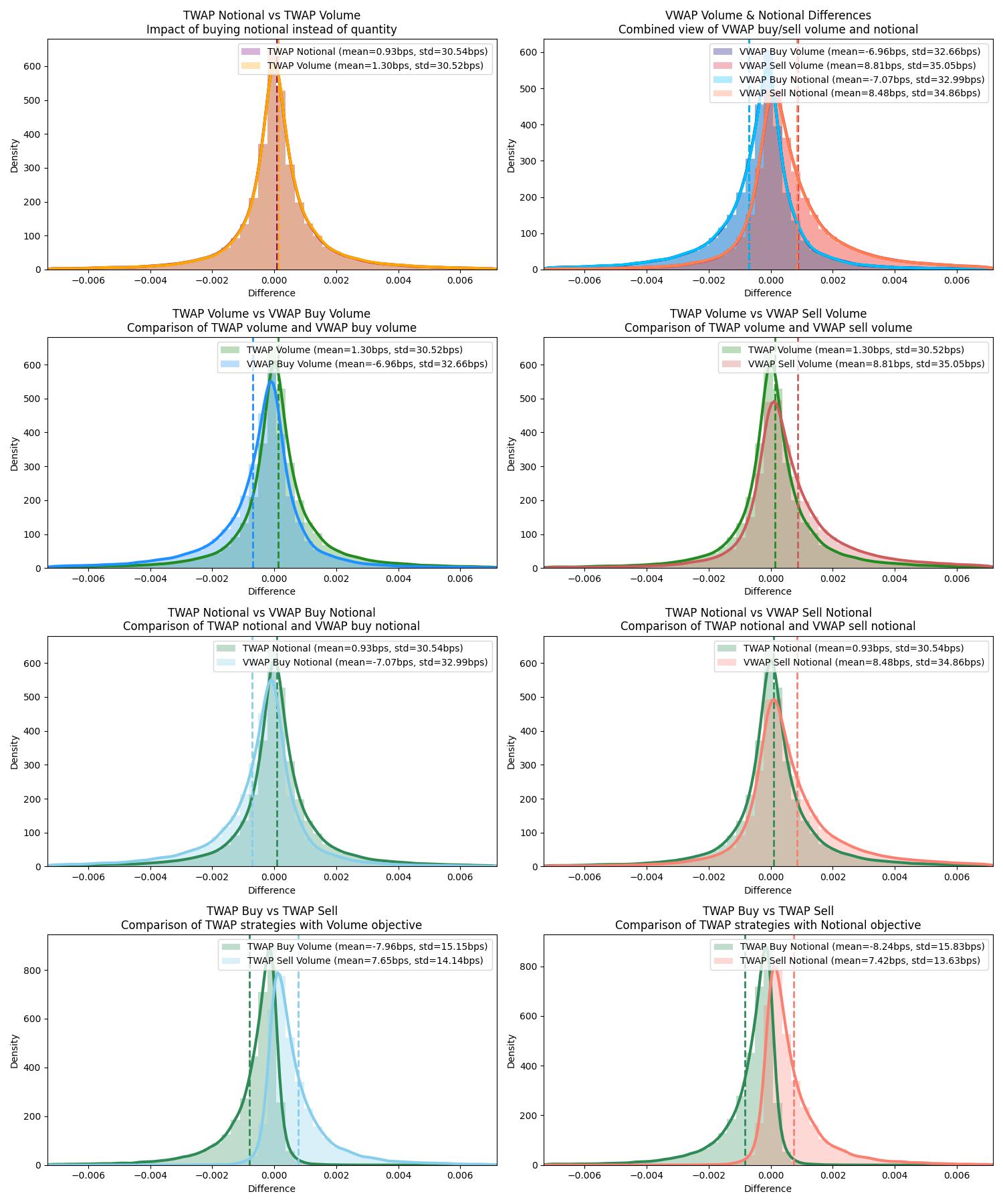}
  \caption{Crypto-currencies - Slippage Distributio - Min 7 periods - Frequency of 15 minutes}
    \label{fig:crypto_6_15}
\end{figure}

\begin{figure}[H]
    \centering
 \includegraphics[scale=0.4]{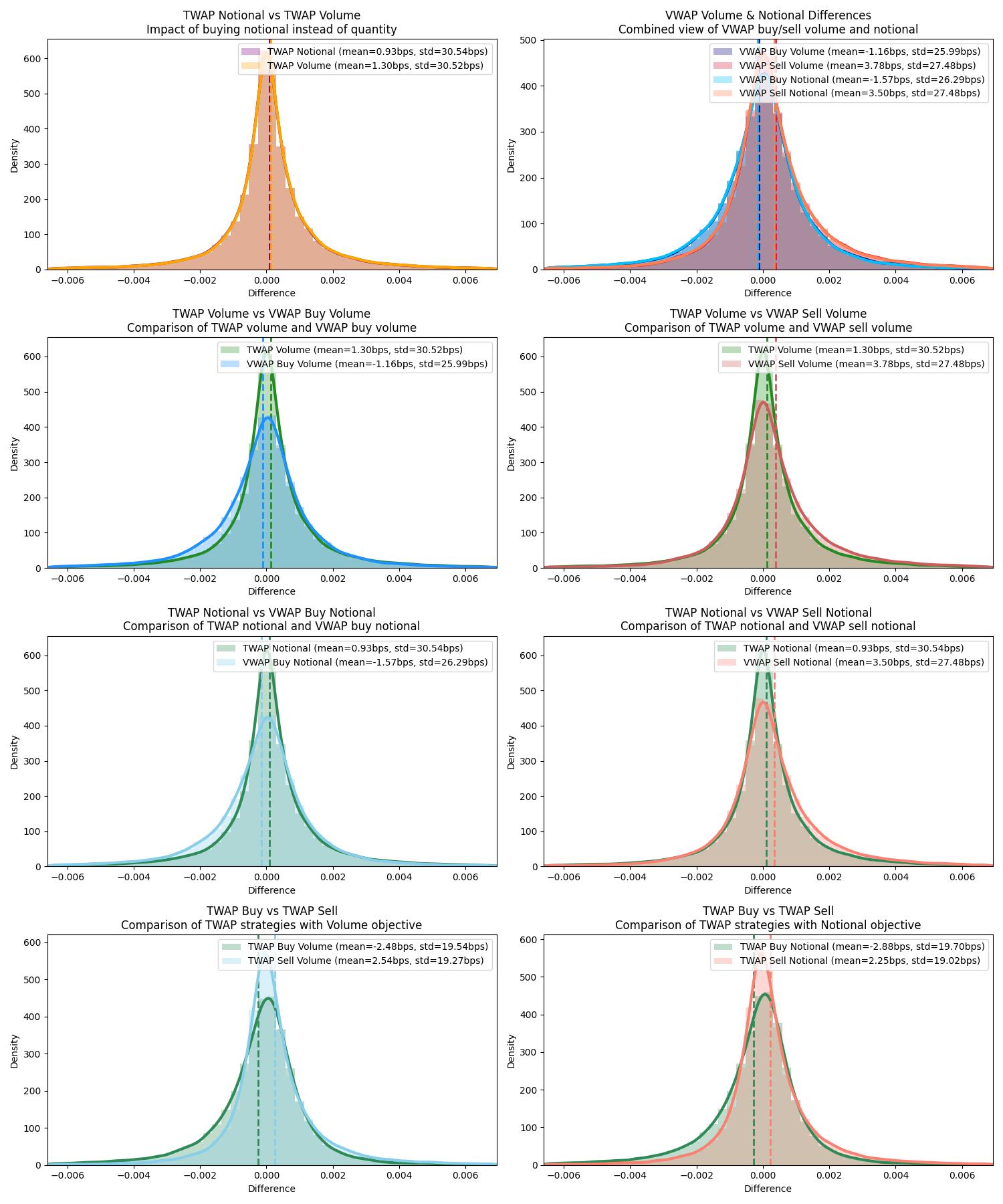}
\caption{Crypto-currencies - Slippage Distribution - Min 12 periods - Frequency of 15 minutes}
    \label{fig:crypto_11_15}
    
\end{figure}

Figures \ref{fig:crypto_6_15} and \ref{fig:crypto_11_15} clearly illustrate the results observed in the tables, with distinctly shifted curves where the losing tail density is consistently dominated by the benchmark TWAP curve, indicating that the results are not driven by extreme positive events but rather by a genuine shift in the distribution.

\begin{figure}[H]
    \centering
 \includegraphics[scale=0.4]{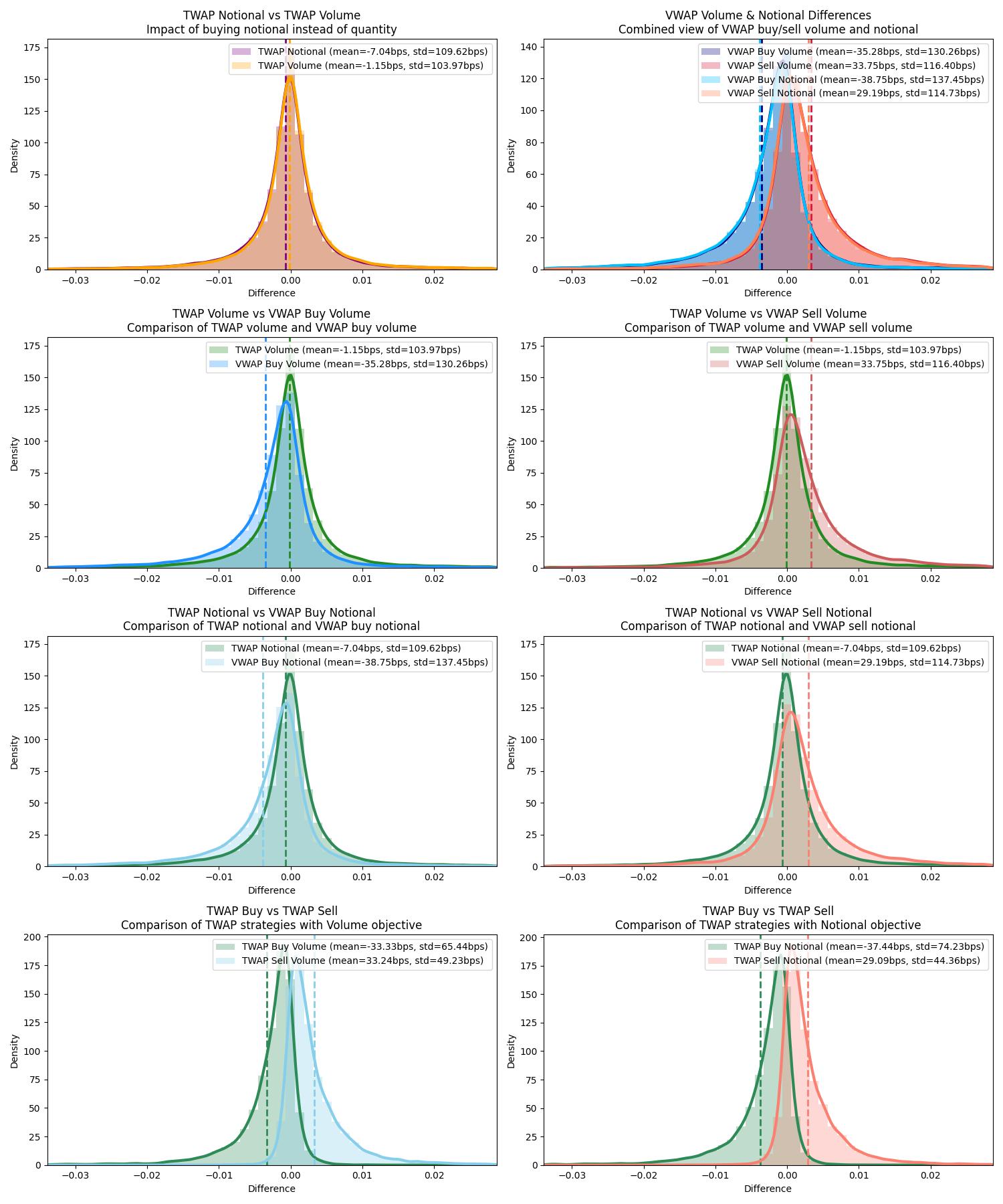}
  \caption{Crypto-currencies - Slippage Distribution - Min 7 periods- Frequency of 250 minutes}
    \label{fig:crypto_6_250}
\end{figure}

\begin{figure}[H]
    \centering
 \includegraphics[scale=0.4]{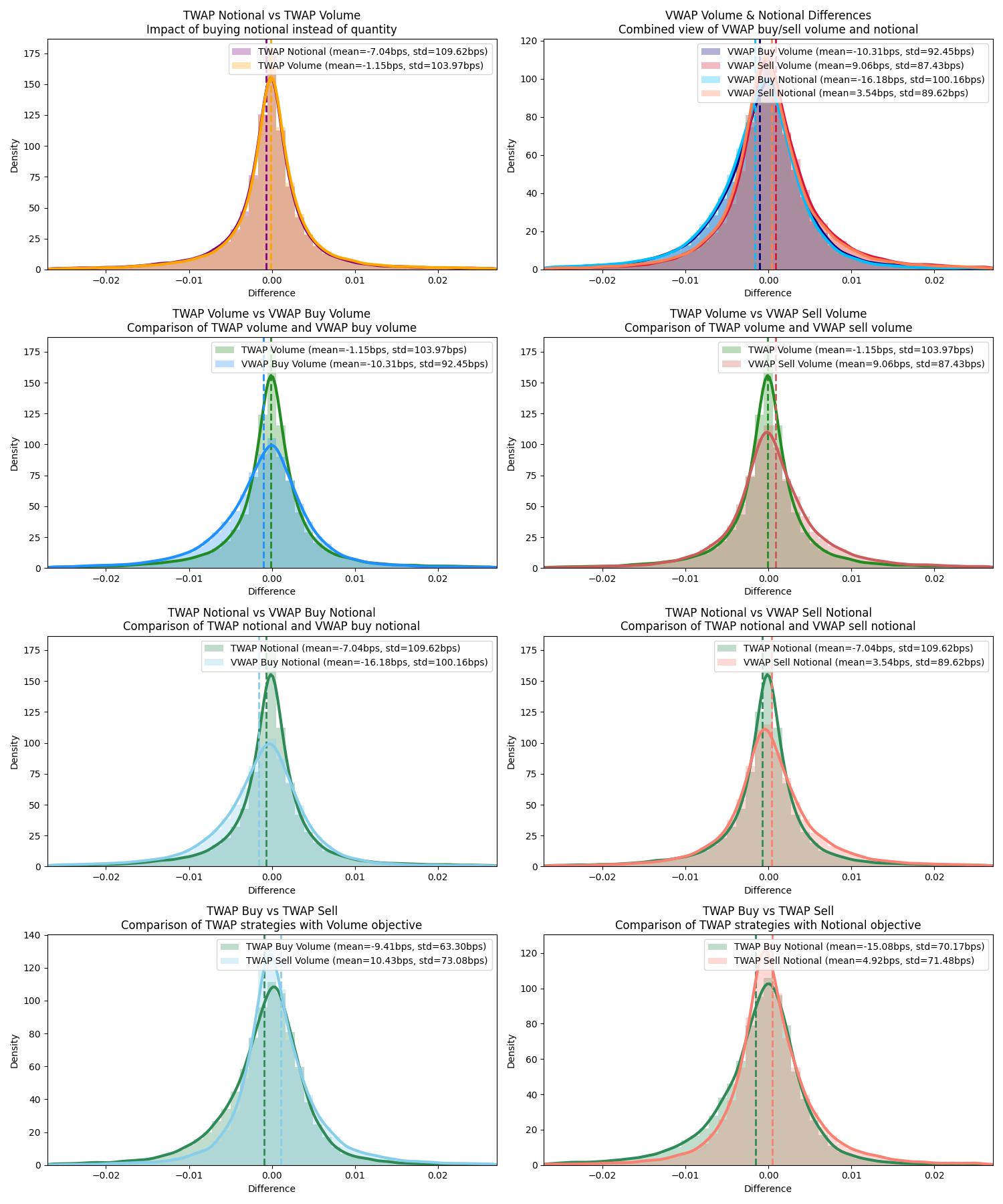}
\caption{Crypto-currencies - Slippage Distribution - Min 12 periods - Frequency of 250 minutes}
    \label{fig:crypto_11_250}
    
\end{figure}

Figures \ref{fig:crypto_6_250} and \ref{fig:crypto_11_250} confirm that the model performs well not only at 15-minute frequency but presents the same slippage distribution shifts across all tested frequencies. Impressively, the TWAP curves even show distribution curves that are nearly disjoint between buy and sell sides. However, it is important to remember that these slippages are measured relative to the benchmark, which is not necessarily constant as its computation depends on when the execution ends. This means that these results do not indicate that the model could be used to directly generate alpha, but rather that it produces alpha in the context of a specific order contract.

\subsection{Understanding the Performance}

To understand the model's behavior, we propose examining the cumulative execution curves in this section. These curves represent the total proportion of an order traded up to time $t$. The figures aggregate all execution curves on the test set for the different order types handled by the model.

\begin{figure}[H]
    \centering
 \includegraphics[scale=0.4]{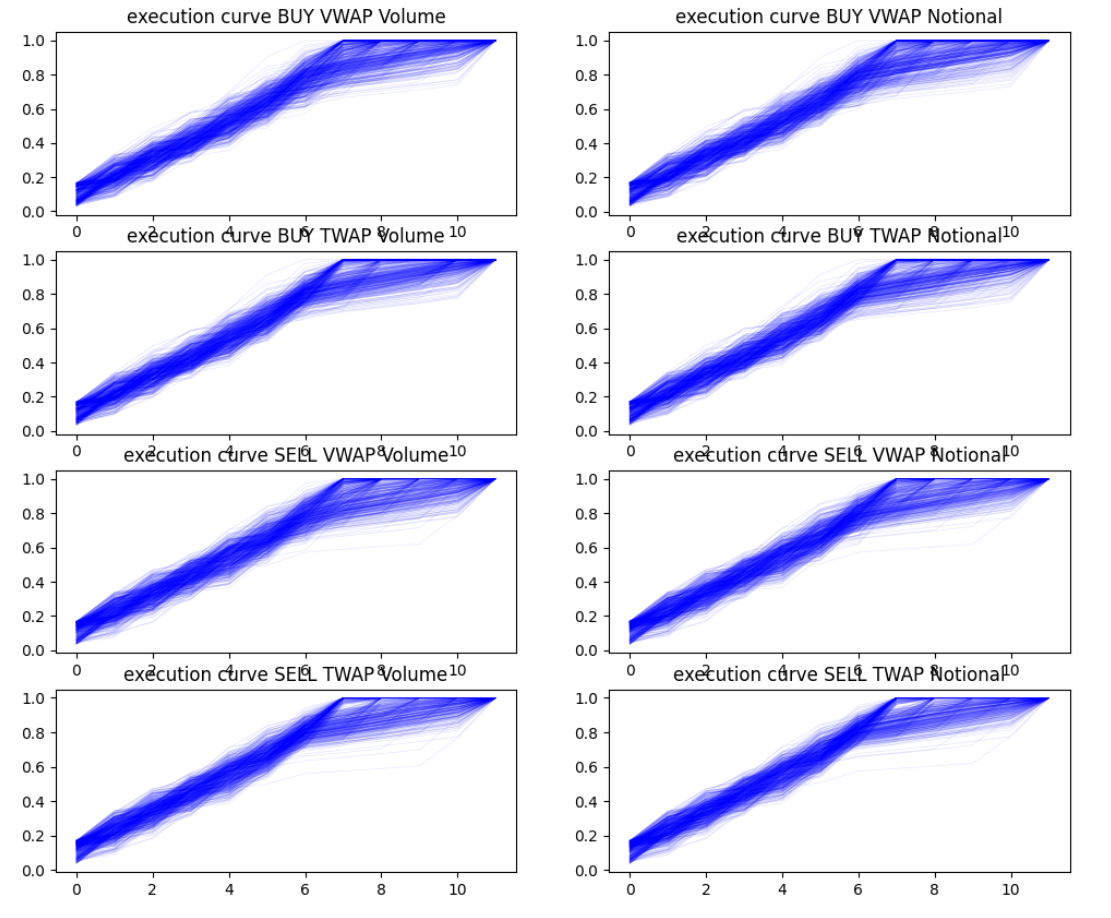}
\caption{Crypto-currencies - Execution Curves - Min 7 periods - Frequency of 15 minutes}
    \label{fig:crypto_exec_6_15}
\end{figure}

Figure \ref{fig:crypto_exec_6_15} shows the case when time flexibility is half the order's maximum duration. In this case, we can clearly see that a large proportion of the execution curves finish earlier than the full execution duration, with orders ending at different time intervals. Quite interestingly, the TWAP and VWAP-based execution curves do not differ substantially when examined closely, but what is observable is that there appear to be more sell orders that wait longer than buy orders. Indeed, this observation makes perfect sense, as the market has grown on average, and the model has simply learned the mean return of the market, which indicates that on average it is better to buy early and sell late.

\begin{figure}[H]
    \centering
 \includegraphics[scale=0.4]{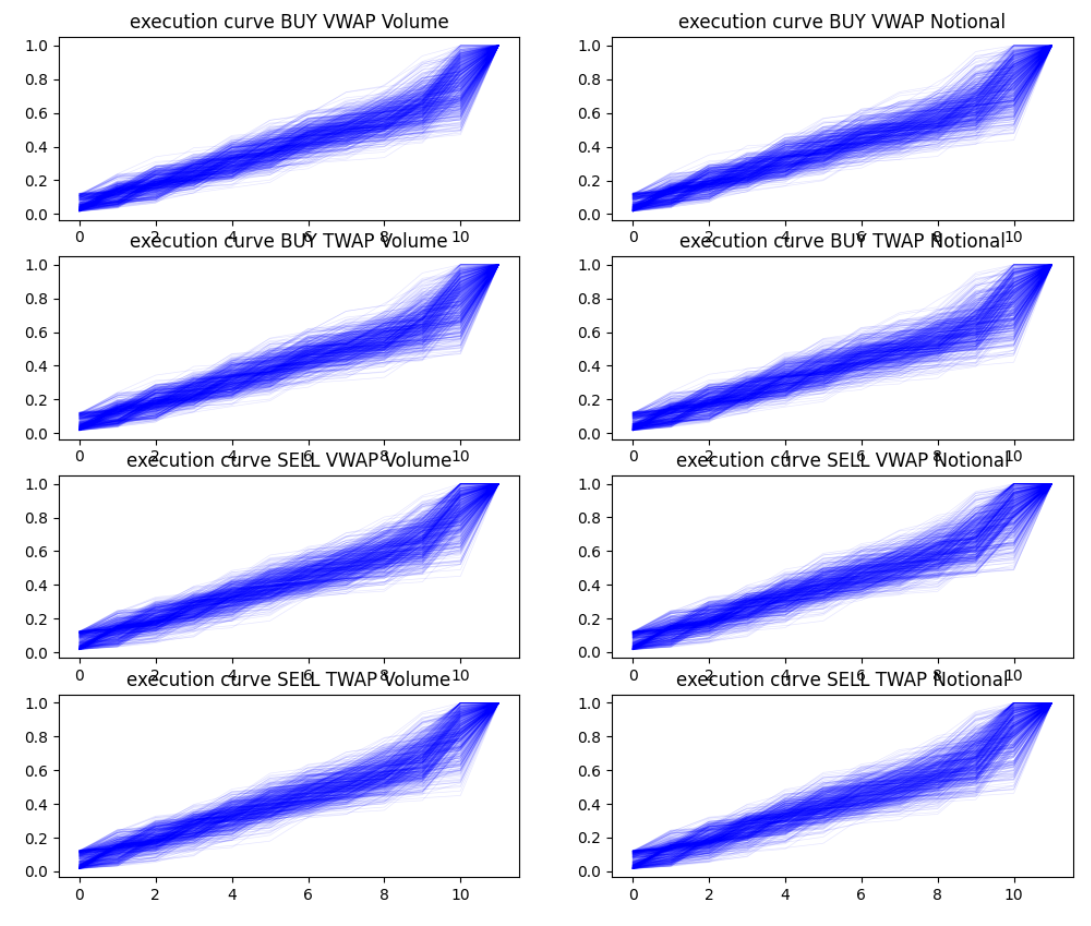}
\caption{Crypto-currencies - Execution Curves - Min 12 periods - Frequency of 15 minutes}
    \label{fig:crypto_exec_11_15}
\end{figure}

Figure \ref{fig:crypto_exec_11_15} presents the case where there is no execution flexibility while the goal is still to beat the benchmark. Here, the model no longer finishes its orders early, indicating that the separated components of the fused-MLP indeed learn very distinct patterns.

\begin{figure}[H]
    \centering
 \includegraphics[scale=0.4]{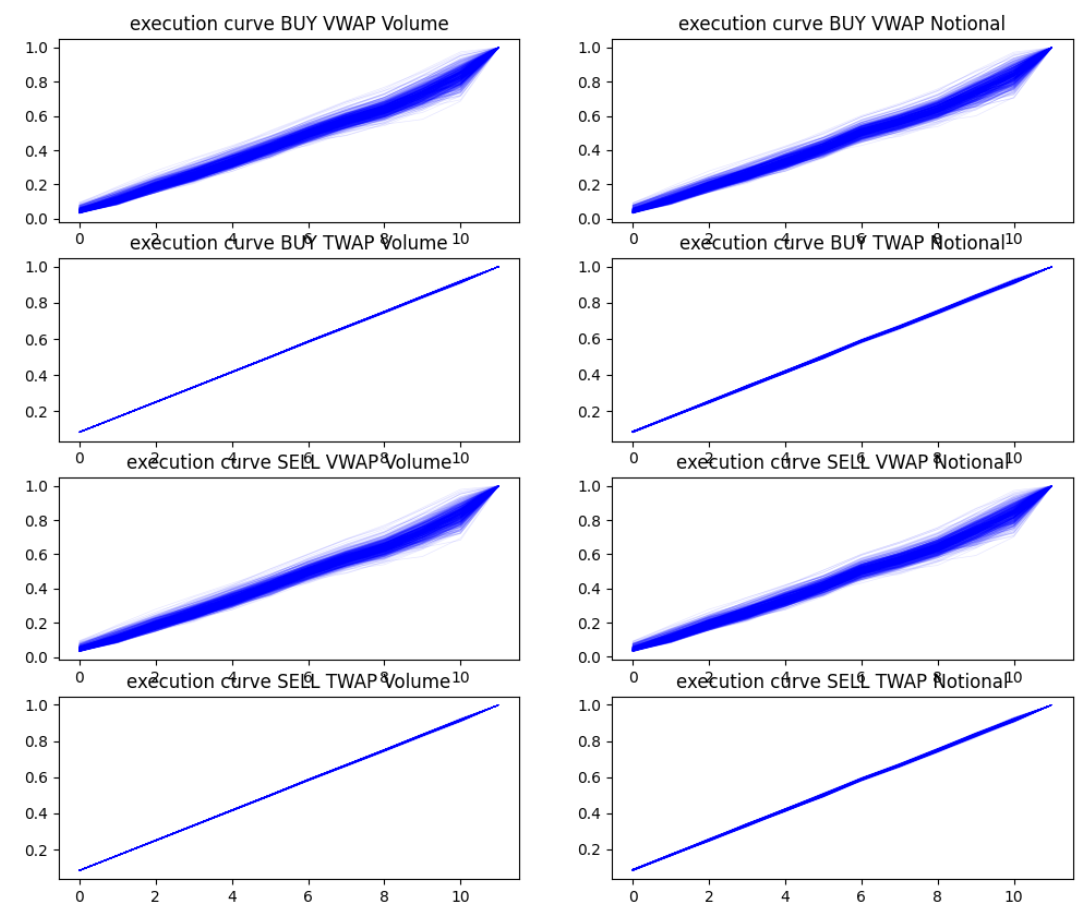}
\caption{Crypto-currencies - Execution Curves - Match VWAP - Frequency of 15 minutes}
    \label{fig:crypto_exec_12_15}
\end{figure}

Finally, Figure \ref{fig:crypto_exec_12_15} shows that models aimed at reducing VWAP risk only have very different execution curves. Indeed, the order of variation between the curves is much slower. We can still observe the distinguishable end-loaded effect previously discussed in \cite{genet2025staticvwap, genet2024dynamic}. Additionally, one can understand why the model matching TWAP does not perform perfectly: while the model is able to learn to approximate a nearly perfect line, this approximation is still not entirely perfect. Nevertheless, the model was able to learn the optimal pattern quite well, even if not completely perfect. However, this component serves more as an example for data structure coherency rather than real practical use.

\begin{figure}[H]
    \centering
 \includegraphics[scale=0.4]{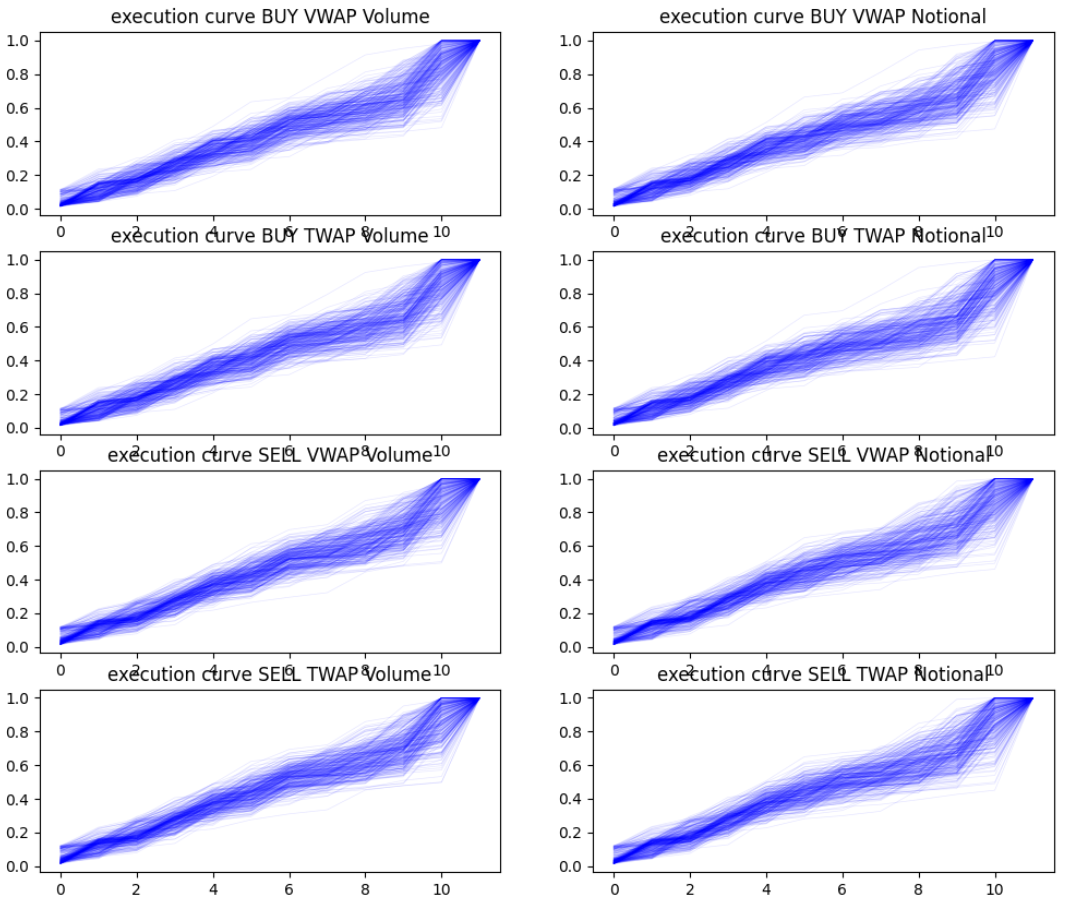}
\caption{Crypto-currencies - Execution Curves - Match VWAP - Frequency of 250 minutes}
    \label{fig:crypto_exec_11_250}
\end{figure}

Quite interestingly, Figure \ref{fig:crypto_exec_11_250} shows that patterns tend to be similar across frequencies, with curves that look similar to those observed at 15-minute frequency, again confirming the viability of merging different frequencies.

\begin{figure}[H]
    \centering
 \includegraphics[scale=0.4]{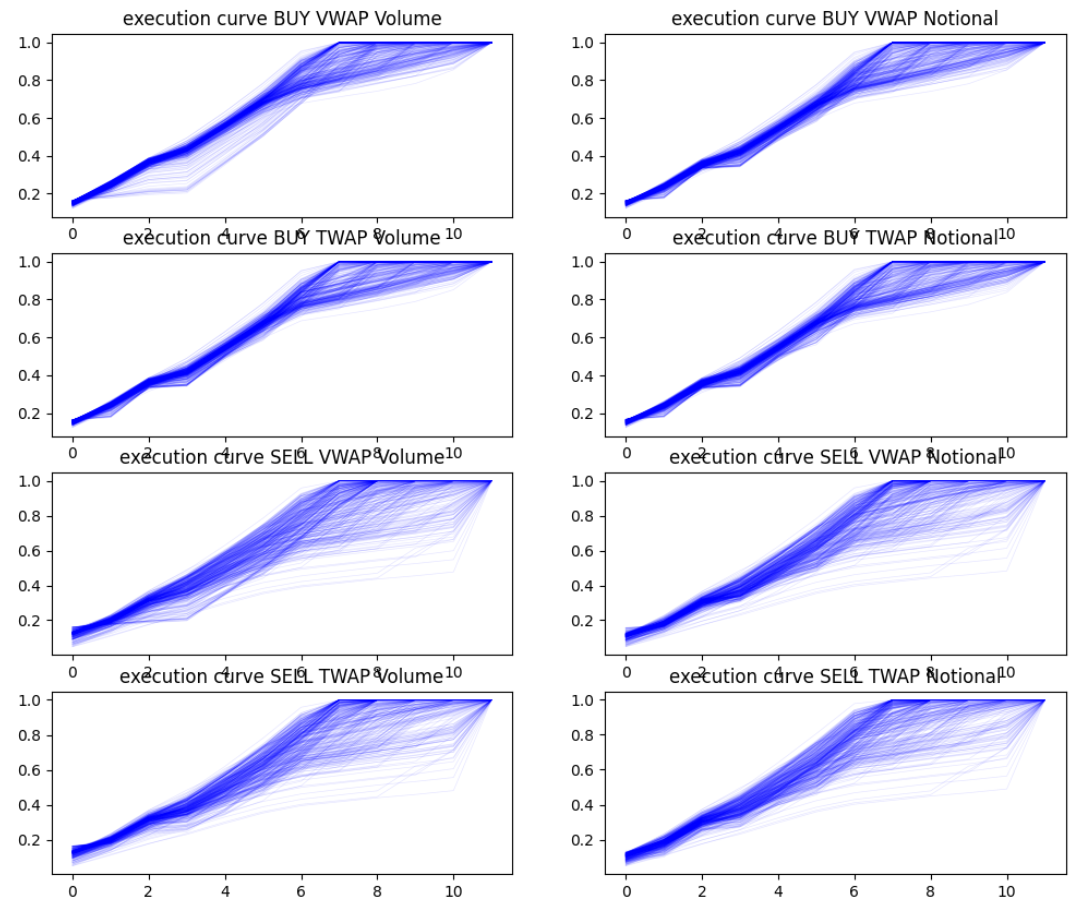}
\caption{Dow Jones - Execution Curves - Min 7 periods}
    \label{fig:dow_exec_6}
\end{figure}

\begin{figure}[H]
    \centering
 \includegraphics[scale=0.4]{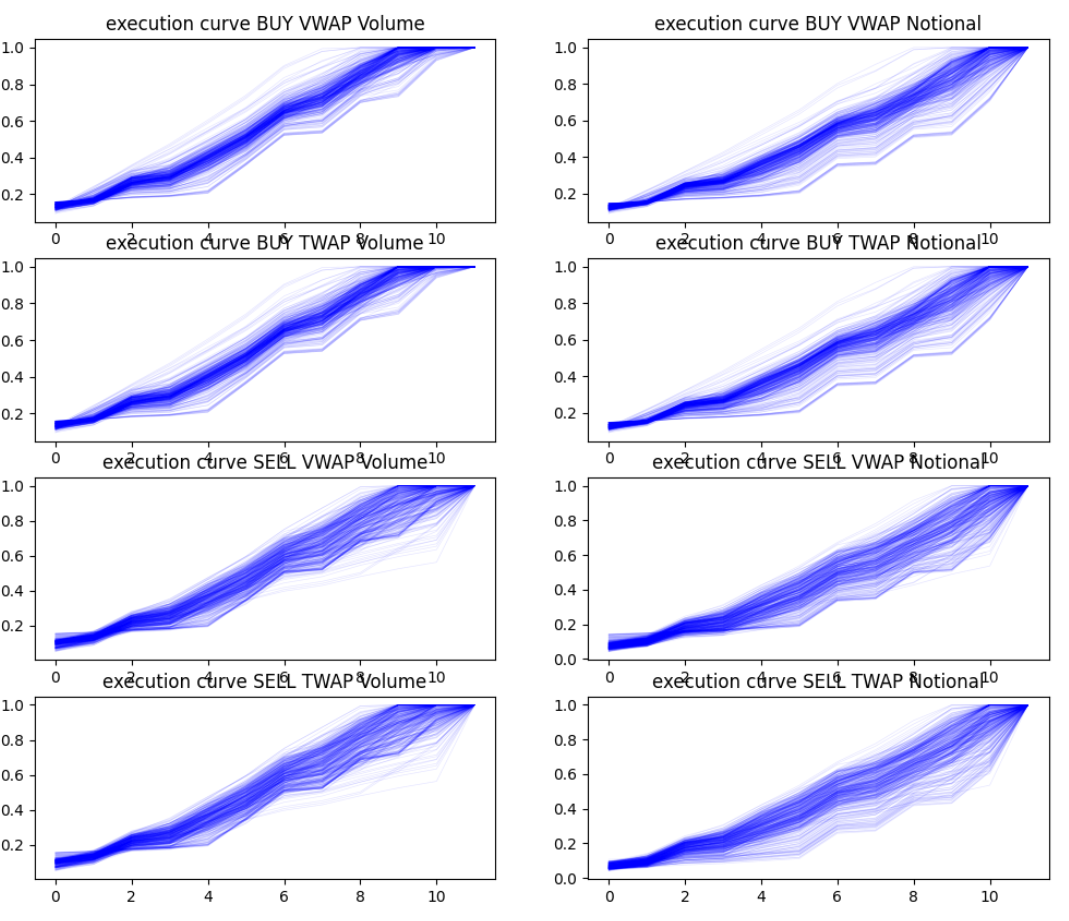}
\caption{Dow Jones - Execution Curves - Min 12 periods}
    \label{fig:dow_exec_11}
\end{figure}

\begin{figure}[H]
    \centering
 \includegraphics[scale=0.4]{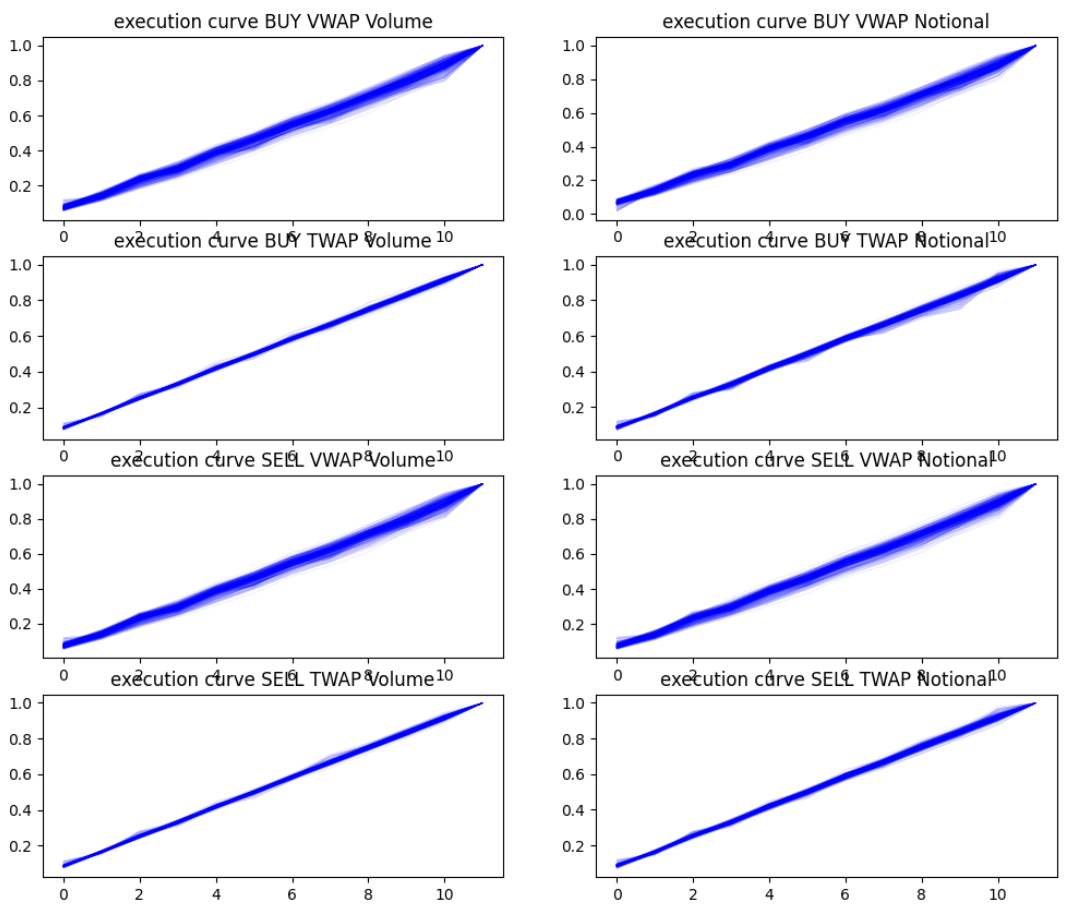}
\caption{Dow Jones - Execution Curves - Match VWAP}
    \label{fig:dow_exec_12}
\end{figure}

Finally, the last three figures \ref{fig:dow_exec_6}, \ref{fig:dow_exec_11}, and \ref{fig:dow_exec_12} present the same curves for the stock case. Here, we can observe that the model presents much more variation even between curves, and especially between buy and sell sides, showing that at lower frequency in the stock market, the average positive returns have a much stronger impact than in crypto markets. It is indeed very interesting to note that in the stock case, even when there is no flexibility, the model component learns to finish earlier, following the general principle that buying early is better than buying late in the stock market on average. Finally, in the last case of matching VWAP, we can observe a similar pattern as for crypto, with much more contained variability in the execution curves and a very slight end-loading effect.

\section{Conclusion}
This paper introduces Large Execution Models (LEMs), a unified deep learning framework for execution problems that enables learning models that work across many assets, frequencies, objectives, and constraints. LEMs decouple market information processing from execution decisions through a shared feature extraction pipeline using TKANs, VSNs, and attention mechanisms, while independent networks handle specific execution logic.

\medskip

The two experiments show that this approach can be implemented consistently across different asset classes, here stocks and cryptocurrencies, and highlight that deep learning's ability to create high-dimensional functions with problem-specific architectures allows easy calibration across different asset classes. In the same way that MLPs are general function approximators, the proposed approach demonstrates an effective execution problem solver.

\medskip

One key result is how execution time flexibility enables obtaining significantly shifted distributions of slippage, confirming the intuition that more freedom in the execution process makes it possible to beat the benchmark. However, one shouldn't be fooled by these results into thinking this represents direct alpha, as the slippage is measured against a benchmark we influence, which differs between order sides, thus making the use of orders to generate direct profits illusory.

\medskip

However, while not generating direct alpha, the model enables easy construction of execution contracts such as share buybacks, providing an all-in-one strategy to maximize profits in such contracts. In addition, LEMs provide significant operational advantages by handling multiple execution scenarios (buy/sell, volume/notional, different time constraints) through a single model deployment. This eliminates the need for asset-specific models and reduces implementation complexity for institutional trading desks. LEMs naturally enable moving non-vanilla VWAP orders with varying parameters from high-touch to low-touch operations.

\medskip

The success with near-vanilla VWAP variations suggests straightforward extensions to more complex contracts with structured execution mandates. Future work could incorporate market microstructure features, multi-asset strategies, or alternative data sources. In conclusion, LEMs confirm the results of \cite{genet2025sigtransvwap} that deep learning is a promising framework for modeling and calibrating execution models, and extend these results by covering more complex cases.
\bibliography{bib}

\end{document}